# Text-to-Image Models and Their Representation of People from Different Nationalities Engaging in Activities

Abdulkareem Alsudais
Prince Sattam bin Abdulaziz University, Saudi Arabia
A.Alsudais@psau.edu.sa

**Abstract**

This paper investigates how popular text-to-image (T2I) models, DALL-E 3 and Gemini 3 Pro Preview, depict people from 206 nationalities when prompted to generate images of individuals engaging in common everyday activities. Five scenarios were developed, and 2,060 images were generated using input prompts that specified nationalities across five activities. When aggregating across activities and models, results showed that 28.4% of the images depicted individuals wearing traditional attire, including attire that is impractical for the specified activities in several cases. This pattern was statistically significantly associated with regions, with the Middle East & North Africa and Sub-Saharan Africa disproportionately affected, and was also associated with World Bank income groups. Similar region- and income-linked patterns were observed for images labeled as depicting impractical attire in two athletics-related activities. To assess image–text alignment, CLIP, ALIGN, and GPT-4.1 mini were used to score 9,270 image–prompt pairs. Images labeled as featuring traditional attire received statistically significantly higher alignment scores when prompts included country names, and this pattern weakened or reversed when country names were removed. Revised prompt analysis showed that one model frequently inserted the word "traditional" (50.3% for traditional-labeled images vs. 16.6% otherwise). These results indicate that these representational patterns can be shaped by several components of the pipeline, including image generator, evaluation models, and prompt revision.

**Keywords:** Text-to-Image; AI ethics; Image generation;

## 1. Introduction

Recent work on Text-to-Image (T2I) models has highlighted various limitations and issues inherent in these systems, with several recent efforts emphasizing systematic biases, evaluation shortcomings, and reasoning errors (Cho et al., 2023; D'Incà et al., 2024; Masrourisaadat et al., 2024; Vice et al., 2025; Zhang et al., 2023). While researchers have examined multiple dimensions of concern, the representation of different nationalities has received less attention. Several studies have revealed concerning patterns where these models generate unfavorable or stereotypical depictions of certain nationalities, cultures, and geographic regions (Bianchi et al., 2023; Dehdashtian et al., 2025; Hall et al., 2023; Jha et al., 2024; Khanuja et al., 2024). According to a survey on T2I biases, current research disproportionately addresses gender and skin tone biases while neglecting geo-cultural representation issues (Wan et al., 2024), with another study noting "considerable room for models to generate more geographically representative content" (Basu et al., 2023). Building on these insights, the primary objective of this paper is to examine a specific aspect of T2I performance: how these systems represent individuals from various nationalities. Specifically, when prompted to generate images describing common everyday activities (e.g., cooking, sports, and drinking coffee at a coffeeshop).

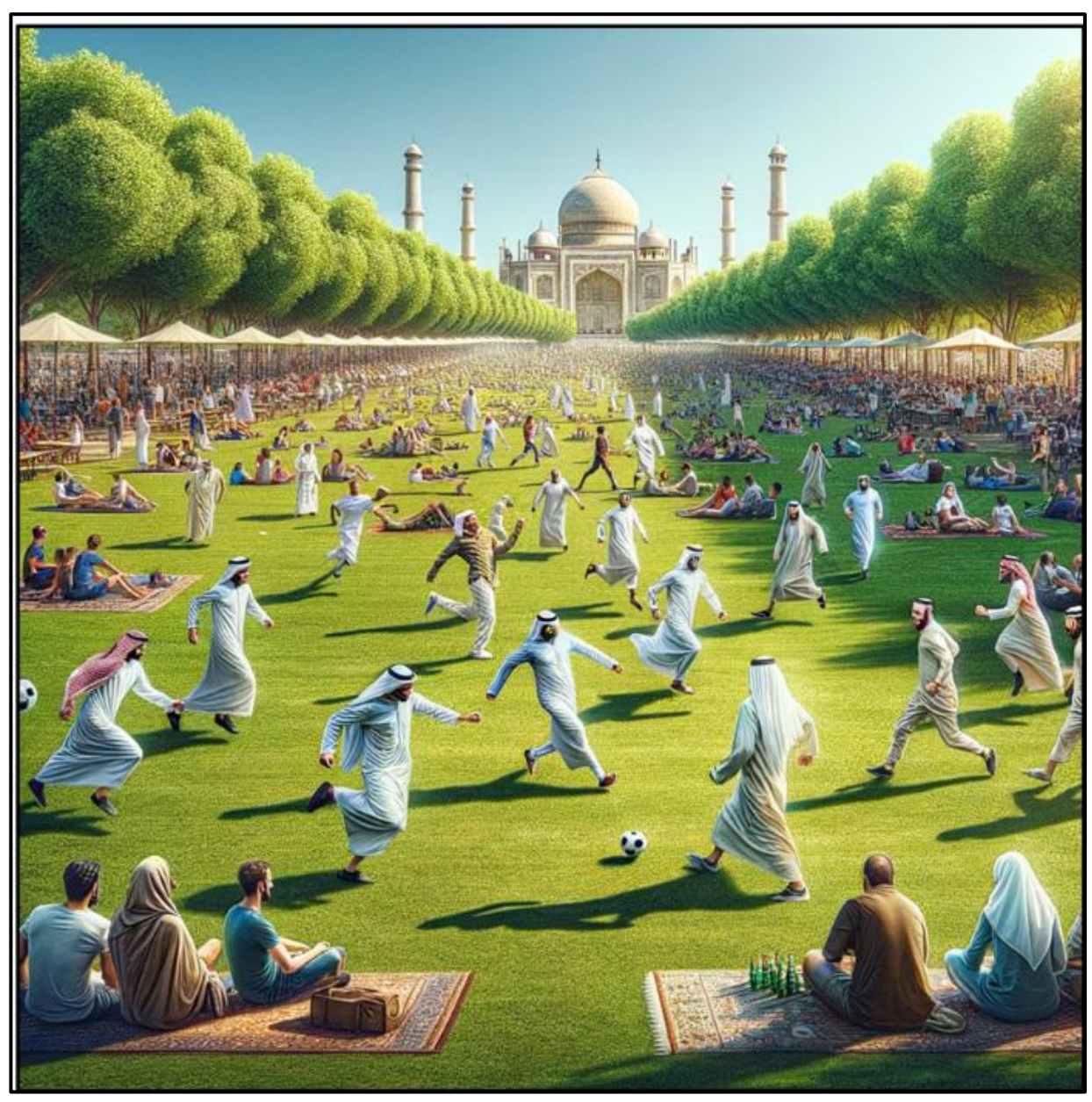

**Figure 1.** Example of an unfavorable representation for people from one country playing soccer, stereotypically depicting them in traditional attire impractical for the activity

This focus is motivated by the observation that T2I and native multimodal models frequently depict individuals from certain regions dressed in traditional attire that aligns with stereotypical global representations, even when such clothing would be impractical for the activities specified in input prompts. Such representations are problematic for several reasons. First, they may not accurately reflect how people from these regions dress in typical settings. Instead, they may reinforce stereotypical views. Second, as T2I and multimodal models become increasingly integrated into various applications, users from misrepresented countries may encounter inaccurate or unfavorable depictions of themselves, potentially leading to negative experiences with these services. Third, such representations may perpetuate stereotypical imagery that could contribute to broader harmful misconceptions about how certain nationalities and cultures should be represented. Therefore, highlighting these issues could help mitigate such unfavorable representations presented in these models.

Previous research in this area has employed various methodologies, including leveraging existing datasets (Askari Hemmat et al., 2024; Bai et al., 2024), gathering annotations from individuals across different countries (Hall, Bell, et al., 2024; Senthilkumar et al., 2024), or focusing on how objects or artifacts are represented (Hall et al., 2023; Jeong et al., 2025). To the best of my knowledge, no comprehensive dataset exists that captures people from an extensive list of countries engaging in identical common activities, which would allow for systematic assessment of how AI models represent different nationalities. Therefore, this study constructs such a dataset that includes 206 nationalities across five scenarios: couples cooking at home, groups playing soccer in public parks, a person working on their laptop from home, a group jogging, and a group drinking coffee at a coffeeshop.

The analysis focuses on two key aspects in the generated images: first, determining whether individuals are depicted in traditional outfits, and second, identifying whether people are shown wearing attire that would be impractical for the specified activity. This deliberately narrow analytical scope offers a specific contribution by examining an aspect not thoroughly explored in previous research. This will allow for a systematic cross-national comparison using the standardized list of activities and their images.

A common challenge when examining AI models is the evaluation of generated images. Various evaluation metrics exist, with recent work focused on creating improved or automated solutions (Hall, Mañas, et al., 2024; Singh & Zheng, 2023; Song et al., 2025), many specifically addressing cultural or geographical elements (Dehdashtian et al., 2025; Hall, Bell, et al., 2024). Several approaches focus on measuring alignment between images and text, with CLIP (Radford et al., 2021) being a widely used resource to quantify this alignment. CLIP has been studied in similar contexts in previous research, with one study finding that alignment scores generated using CLIP were higher for images representing high-income groups compared to lower-income ones (Nwatu et al., 2023). The authors in another study discovered that CLIP may favor stereotypical representations for certain groups (Agarwal et al., 2021). Another pre-trained model commonly utilized in vision and language task is ALIGN (Jia et al., 2021). Recent multimodal transformer foundation models can also process images and, following defined instructions, generate scores that measure the alignment between an image and its caption. GPT-4.1 mini, which was released in April 2025 (OpenAI, 2025), can be used in this capacity.

This paper contributes to this line of research by examining CLIP, ALIGN, and GPT-4.1 mini performances in the context of nationality representation, specifically investigating whether there are statistically significant differences in alignment scores between labeled images while testing multiple prompt variations. A final analysis introduced in this paper includes examination of “revised prompts,” which are the enhanced prompts generated by OpenAI when using their API to generate images with DALL-E 3 to add details to the original input prompts provided by users.

These revised prompts are analyzed to identify if elements they introduce may contribute to potentially unfavorable representations of individuals from different countries. It is important to note that this paper does not propose a new approach that outperforms existing state-of-the-art solutions. Instead, the study focuses on examining a specific aspect of text-to-image models to investigate whether evidence exists for representational biases affecting certain nationalities.

In summary, this paper investigates how two T2I models represent people from 206 countries when prompted to generate images specifying nationalities and five typical everyday activities: cooking a meal at home, playing soccer in a public park, a person working on a laptop from home, a group jogging, and a group drinking coffee at a coffeeshop. DALL-E 3 (OpenAI, 2023) and Gemini 3 Pro Preview (Google DeepMind, 2025), which is also referred to as Nano Banana Pro, are used to generate images for these five scenarios across 206 nationalities.

The generated images are manually labeled to indicate whether the depicted individuals are dressed in traditional or impractical attire. These labeled images are then analyzed to examine potential correlations between countries' regions and income groups and how their people are depicted in the generated images. Subsequently, CLIP, ALIGN, and GPT-4.1 mini are used to compare alignment scores to determine whether it assigns higher scores to images labeled as featuring “traditional” or “impractical” attire. Finally, the study investigates revised prompts to examine whether text added to input prompts contributes to how individuals from specific countries are depicted.

## 2. Research Objectives

The main objective of this paper is to study how T2I models represent individuals from various countries when prompted to generate images of people engaging in typical activities. Specifically, the paper focuses on the following three primary research questions, which are explored by generating images using DALL-E 3 and Gemini 3 Pro Preview for five scenarios across 206 nationalities:

- RQ1: When generating images of people from various countries performing common activities, to what extent are they depicted in traditional or impractical attire, and how does this correlate with the country's region or income group?
- RQ2: When employing CLIP, ALIGN, and GPT-4.1 mini to calculate alignment between images and prompts, to what extent do alignment scores correlate with the presence of traditional or impractical attire in labeled images?
- RQ3: When examining revised prompts, what elements do they introduce to the final instructions, and how do these elements contribute to the representation patterns identified in RQ1?

## 3. Related Work

Previous research has examined AI and how it affects various aspects of life (Charef et al., 2023; Xu et al., 2024). For T2I models specifically, several studies have addressed how they unfavorably depict individuals from certain nationalities or cultures (Ghosh et al., 2024; Qadri et al., 2023). Jha et al. (2024) examined visual and non-visual stereotypes affecting people from various cultures in T2I-generated images, finding that these issues disproportionately impact individuals from specific regions. Building on these findings, recent work has developed methods to detect or measure problematic representations (Dehdashtian et al., 2025; Qadri, Diaz, et al., 2025).

Visual language models (VLMs) exhibit similar regional problematic representations. In one paper, the authors demonstrated that one model performs significantly better on questions about images from one region compared to those from other regions, though performance improved when self-correction was employed (Cui et al., 2023). The authors in another paper proposed a new evaluation approach for VLMs that incorporates multiple dimensions including fairness and toxicity (Lee et al., 2024).

Research on cultural representation extends beyond T2I to broader language model contexts (Liu et al., 2024; Pawar et al., 2025). Qadri et al. (2025a) note that “language models will shape how people learn about, perceive and interact with global cultures,” emphasizing the importance of considering “whose knowledge systems and perspectives are represented in models.” The authors in a related study found that LLMs made more mistakes when responding to questions about countries in Sub-Saharan Africa compared to North America (Moayeri et al., 2024). Some researchers have focused on creating datasets that mitigate current limitations affecting particular regions (Bhutani et al., 2024). In the end, these findings point to the need for dedicated efforts in addressing cultural and geographical representation issues in AI models. This paper contributes to this focus by targeting a narrow and specific issue to examine it from multiple angles.

# 4. Methods

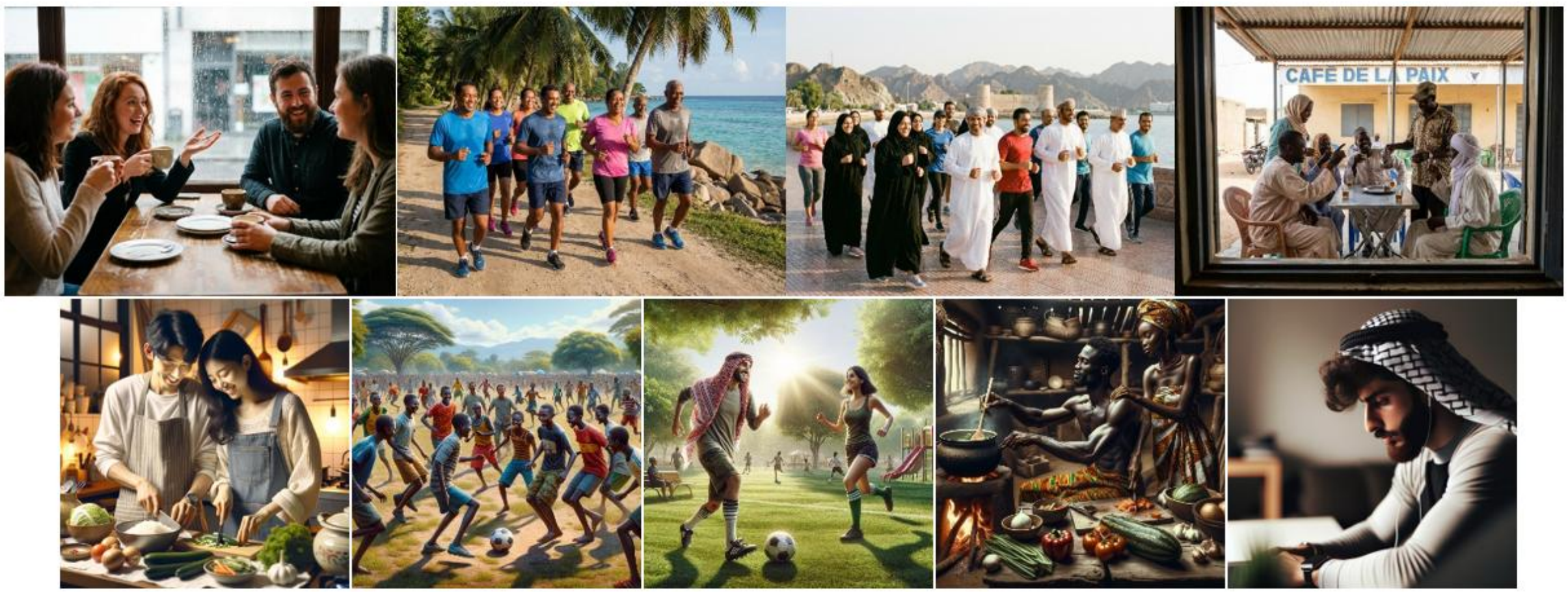

**Figure 2.** Samples of image used in this study

To investigate how T2I models depict individuals from various countries engaged in typical activities, the first step involved establishing a comprehensive nationality list. Using the World Bank classification as the source, countries that had both region and income group designations were selected. This process yielded 209 countries across seven regions and four income categories. For image generation, DALL-E 3 (OpenAI, 2023) and Gemini 3 Pro Preview (Nano Banana Pro) (Google DeepMind, 2025) were used. DALL-E 3 was selected for this analysis because it is a widely used text-to-image model. Gemini 3 Pro Preview was included because it is a recently released model that has reported strong performance on multiple vision-language benchmarks, outperforming many established systems.

The initial analysis in this paper focused on images generated using DALL-E 3; Gemini 3 Pro Preview was subsequently added to assess whether the representation concerns identified extend beyond a single model. In addition, because Gemini 3 Pro Preview is a newer release, its inclusion provides an indication of whether these issues persist in more recent generations of T2I models. Finally, these are two proprietary models accessible through two of the most widely used AI platforms: DALL-E 3 is developed by OpenAI and Gemini 3 Pro Preview by Google. Their inclusion therefore helps ensure that the study evaluates widely used, commercially deployed T2I systems that serve large user bases.

Gemini 3 Pro Preview (Nano Banana Pro) was used through the official Google API while DALL-E 3 was employed through OpenAI's official API. For DALL-E 3, results collected included both the generated images and the API's “revised prompts” for later analysis. During image generation, the OpenAI API used to generate the images raised safety flags whenever “South Sudan” was included in prompts, requiring its exclusion. Subsequent analysis also prevented image generation for “Sudan” and “Myanmar”. This required reducing the final sample to 206 nationalities.

Additional details about the dataset and the number of images collected are in subsection 4.4. Figure 2 presents sample images generated and analyzed in this study. The four images in the top row were generated using Gemini 3 Pro Preview, while the five images in the bottom row were generated using DALL-E 3. For each model, the first two images shown are examples of images not labeled as “traditional” or “impractical,” whereas the remaining images are examples of images labeled as “traditional,” “impractical,” or both. The sample images span the five activities examined in this study.

| Code | Activity | Text |
|---|---|---|
| C1 | Cooking a meal | Generate an image of a couple from [country] cooking a meal in their home |
| D1 | Drinking coffee at a coffeeshop | A candid photograph of a group of people from [country] drinking coffee at a coffeeshop |
| J1 | Jogging | A candid photograph of a group of people from [country] jogging |
| S1 | Playing Soccer | Generate a realistic photograph-like image as if it was taken by a photographer of a group of people from [country] playing soccer in a public park |
| W1 | Working from home on a laptop | A candid, documentary-style photograph of a person from [country] working from home. The person is in profile, deep in thought and looking intently at their laptop screen, completely ignoring the camera |

**Table 1.** List of activities used in this study and the full prompts used to generate images

Five distinct scenarios were designed to analyze how T2I models depict individuals from various nationalities in common activities. The scenarios were designed to capture a broad range of elements and activity categories that reflect everyday life. Specifically, two of the five scenarios took place in home settings, two focused on sports or exercise activities, and one was situated in a public establishment. This mix was intended to support the generalizability of the findings by avoiding an overly narrow focus on a limited set of activities. Instead, the scenarios represent a range of activities that people commonly perform across different contexts and cultures, span both indoor and outdoor settings, include both individual and group depictions, and allow for clear labeling in the context of this study.

The first two activities examined domestic settings. The first relied on using the prompt: "Generate an image for a couple from [country] cooking a meal in their home." Each resulting image was analyzed to determine whether at least one person appeared in traditional attire. Similarly, the second prompt specified a person working from home on a laptop (W1 in Table 1). The third and fourth scenarios focused on recreational activities using the prompts: "Generate a realistic photograph-like image as if it was taken by a photographer of a group of people from [country] playing soccer in a public park" and "A candid photograph of a group of people from [country] jogging".

For these scenarios, images were evaluated on two categories: 1) whether they featured people in traditional clothing and 2) whether the depicted attire was impractical for playing soccer or jogging such as formal wear, cultural clothing, or other similar apparel. A final scenario examined public settings by specifying an image of people drinking coffee at a coffeeshop. The exact prompt used was "A candid photograph of a group of people from [country] drinking coffee at a coffeeshop." The complete prompts used to generate images with the two models are provided in Table 1. Each prompt is assigned a unique code, and revised prompts are designated with a second code for the relevant scenarios (i.e., C2, S2, and J2).

These two labels ("traditional" and "impractical") were selected as the primary representational indicators in this study because they capture two distinct aspects of how people from different nationalities are depicted in generated images. First, the "traditional" label captures the tendency for models to default to traditional or culturally marked representations that emphasize specific forms of attire. Second, the "impractical" label captures a functional mismatch between the requested activity and the depicted clothing, particularly for athletics-related activities where the suitability of attire for the activity context can be assessed straightforwardly.

## 4.1 RQ1: Investigating Representation in Generated Images

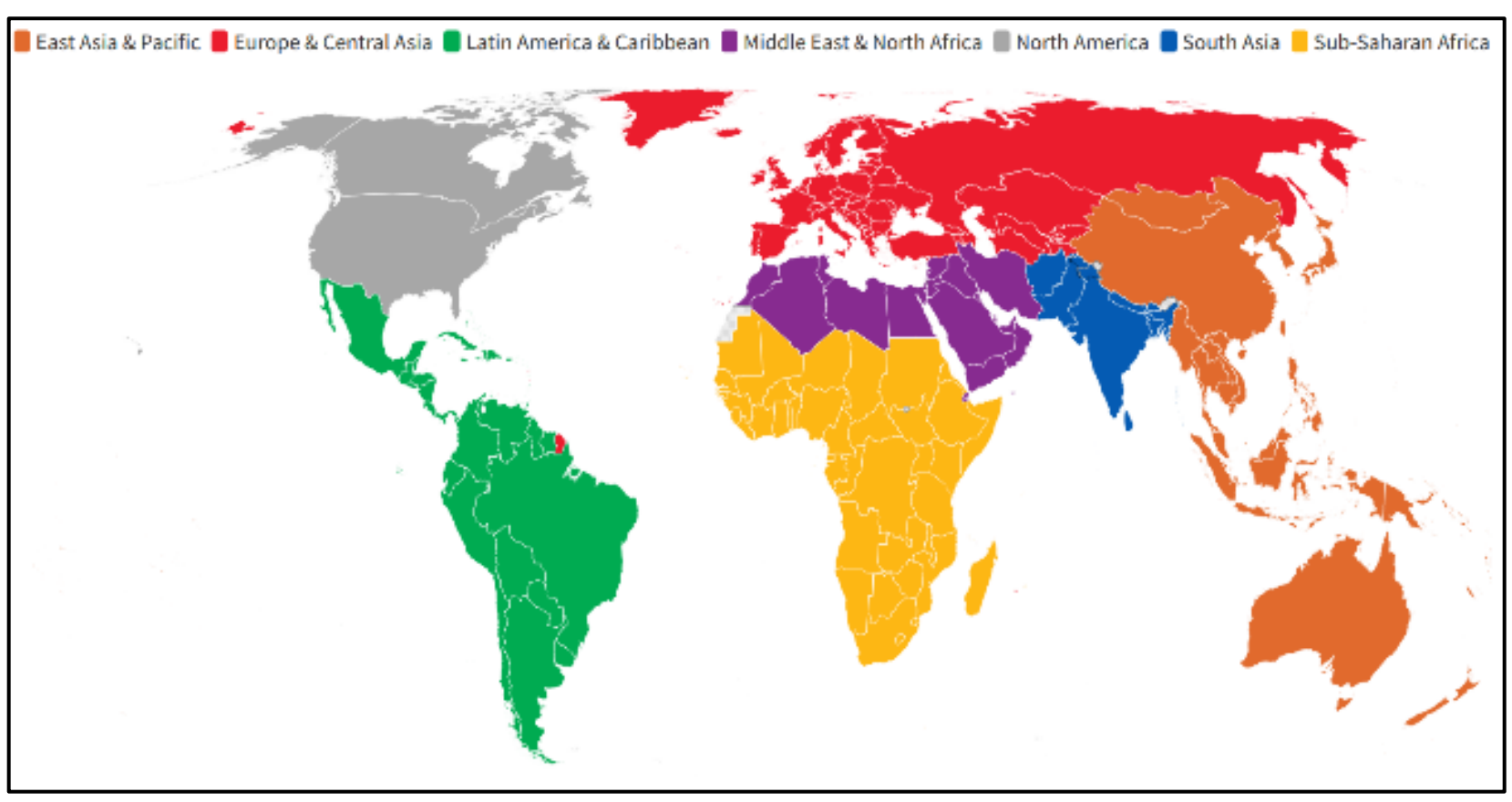

**Figure 3.** Classification of regions according to the World Bank. Source of image: World Bank

The first research question starts by quantifying instances where the T2I model generated images of individuals wearing traditional outfits. To address this, all generated images from the five activities and the two models were qualitatively labeled to identify traditional clothing. Rather than attempting to determine officially recognized traditional outfits for each country, the annotation process simply marked whether an outfit appeared to be traditional. Clothing strongly associated with the broader region in which the country is located was also labeled as "traditional." For example, an image for a country showing individuals in attire similar to traditional clothing from a neighboring country was still labeled as containing traditional outfits. This approach acknowledged the challenge of defining "official" traditional clothing for each nationality and that neighboring countries often share cultural traits, just as dialects often overlap, as discovered for Arabic (Alsudais et al., 2022). This is also motivated by the observation that images generated for countries experiencing current or recent conflicts appeared to undergo safety filtering that defaulted to regional representations. This may be due to the dataset used to train the models containing many war related images, similar to how the original ImageNet included many war related images under the category "Iraqi" (Alsudais, 2022).

For soccer and jogging images, two distinct assessments were conducted. First, images were evaluated for clothing impracticality. In this case, any person wearing attire unsuitable for the activity resulted in a "impractical" label. Second, the presence of traditional outfits among either players or spectators was assessed, with images labeled as "traditional" when this condition was met. The labeling approaches differed slightly between these two categories. For the impracticality, the primary focus was on attire appropriateness for athletic activity, and formal wear like suits was marked as "impractical." When only spectators (and not participants) wore traditional clothing, the image was not labeled impractical but could be labeled "traditional" in the category tracking traditional attire. A single image could therefore be labeled both "impractical" and "traditional".

Following this labeling process, the first part of RQ1 is addressed by summarizing the results obtained from applying it to all images generated using the two models across the five scenarios. In total, the five scenarios were assessed twice for the "traditional" category, resulting in 10 assessments per dataset when considering both models. Additionally, the "impractical" category was examined for the soccer and jogging scenarios for both DALL-E 3 and Gemini 3 Pro Preview, resulting in four additional sets of results. Finally, results were aggregated by activity (cooking, coffee, jogging, soccer, and working), model (DALL-E 3 and Gemini 3 Pro Preview), and label category ("traditional" and "impractical"), resulting in 13 additional sets of results.

To analyze regional patterns, countries were aggregated according to the World Bank regional classifications: East Asia & Pacific, Europe & Central Asia, Latin America & Caribbean, Middle East & North Africa, North America, South Asia, and Sub-Saharan Africa (Figure 3). Results were calculated to show the percentage of images for each region across all sets of results. Chi-square tests of independence were employed to determine whether statistically significant associations existed between regional classifications and these categories. A parallel analysis was conducted using the World Bank's income group classifications: Low Income, Lower Middle Income, Upper Middle Income, and High Income, applying the same statistical testing to identify potential correlations between economic status and representation patterns. To quantify effect size, Cramér's V was used to measure the strength of the association.

## 4.2 RQ2: Exploring CLIP, ALIGN, and GPT-4.1 mini

The second research question examines the effectiveness of CLIP (Radford et al., 2021), ALIGN (Jia et al., 2021), and GPT-4.1 mini (OpenAI, 2025) for evaluating alignment between prompts and generated images, particularly when prompts include country or nationality references for human representation. A motivating hypothesis is that these two pretrained vision-language models, CLIP and ALIGN, while valuable in other contexts, may be inadequate for assessing generated images when nationalities are specified in people-centered prompts.

CLIP and ALIGN were trained on millions of image-text pairs and works by projecting both text and images into a shared embedding space. Accordingly, the cosine similarity between these embeddings can be calculated and used to assess alignment, with perfect alignment resulting in a maximum value of 1. GPT-4.1 mini is a multimodal transformer foundation model that was released in April 2025. It represents one of OpenAI's most recent releases and it combines standard large language model capabilities with advanced multimodal features, including the capability to perform various vision-language tasks. Notably, the model can process images and evaluate their semantic alignment with corresponding textual prompts.

CLIP, ALIGN, and GPT-4.1 mini were selected for three primary reasons. First, CLIP and ALIGN represent widely adopted pretrained vision-language models in the field. Given that one objective of this paper is to assess how commonly employed models evaluate nationality representation in generated images, utilizing CLIP and ALIGN allow for determining whether established models can adequately capture the issues highlighted in this study. Second, GPT-4.1 mini overcomes traditional limitations of text-only GPT models by incorporating multimodal capabilities. It was selected as it represents an emerging approach to vision-language tasks that is likely to be integrated into diverse applications. Another primary motivation for its inclusion stems from its recent release. If even this advanced model fails to identify the issues highlighted in this paper, this would suggest that the challenges documented here persist across the current generation of vision-language models. CLIP was accessed through Hugging Face's Transformers library (Wolf et al., 2020) using the "OpenAI/clip-vit-base-patch32" (ViT-B/32) implementation. For ALIGN, while originally developed by Google, no official public implementation is currently available. Consequently, an open-source version trained on the COYO-700M dataset was utilized via the same Hugging Face library. This implementation, “kakaobrain/align-base”, has been used in several published studies (Mandal et al., 2024; Nakata et al., 2025). GPT-4.1 mini was accessed through the official OpenAI API. Both CLIP and ALIGN can directly compute alignment scores between text and images. For GPT-4.1 mini, specific instructions were provided when accessing it using the API to clearly define the task’s objective. The standard ChatGPT interface and the model “o3” were used to help develop the prompt that was incorporated into the API calls. The prompt begins as follows, with the complete version provided in the appendix:

```
### Goal
Given **one image** and **the following caption**:{caption_text}", you must:
1. Compute an **Alignment Score** ∈ [-1, 1] (floating-point, any precision).
  - **-1.0** → completely unrelated or contradictory
  - ** 0.0** → neutral / no discernible relation
  - ** 1.0** → perfectly matches in all salient aspects
  The score should approximate a cosine similarity between the visual scene and the text.
```

To answer RQ2, alignment values based on CLIP, ALIGN, and GPT-4.1 mini were calculated for each generated image against its corresponding prompt variations. For each scenario, the evaluation included: (1) the original prompt used to generate the images (Table 1), (2) the revised prompt generated by OpenAI's system and provided to DALL-E 3 model, and (3) three modified variations of the original prompt designed to test specific aspects of nationality representation (Table 2). This multi-prompt evaluation approach enables a comprehensive assessment of how different textual specifications influence image-text alignment metrics.

Revised prompts are automatically generated by OpenAI to enhance the input prompts provided for DALL-E 3 image generation. These revised prompts have been studied in one previous research (Jahani et al., 2024). For example, when given the input prompt “generate an image of a couple from Saudi Arabia cooking a meal in their home,” the system generated this more detailed revised prompt: “*an image of a Middle-Eastern couple, wearing traditional Saudi Arabian attire, in their home kitchen. They are standing next to each other, engaged in the process of preparing a traditional Saudi dish, which involves various ingredients spread out on the kitchen counter. The kitchen should be detailed, reflecting the unique architectural style of Saudi homes, with geometric patterns on the fixtures and wooden cabinets. The couple looks happy, sharing fond smiles and engrossed in their culinary task.*”

To calculate alignment values for revised prompts for CLIP and ALIGN, each prompt was first segmented into individual sentences. Values were calculated for each sentence separately, then aggregated to generate an overall alignment metric for the complete revised prompt. This approach was followed due to the length of revised prompts, which consisted of four or five sentences. After examination of a sample of revised prompts, it was observed that each sentence contained comparable amounts of visual information, with all sentences describing relevant visual elements that should appear in the images. Therefore, this approach to calculating alignment scores for revised prompts was determined to be appropriate for this analysis. Calculating scores for the revised prompts was straightforward with GPT-4.1 Mini. Since the model can process the entire prompt as a whole, it was not necessary to evaluate each sentence individually.

| Activity | Code | Text |
|---|---|---|
| Cooking a meal | C3 | A couple from [country] cooking a meal in their home |
| | C4 | A couple from [country] cooking a meal |
| | C5 | A couple cooking a meal in their home |
| Drinking coffee at a coffeeshop | D3 | A group of people from [country] drinking coffee at a coffeeshop |
| | D4 | A group of people from [country] drinking coffee |
| | D5 | A group of people drinking coffee at a coffeeshop |
| Jogging | J3 | A group of people from [country] jogging |
| | J4 | People from [country] jogging |
| | J5 | A group of people jogging |
| Playing soccer | S3 | A group of people from [country] playing soccer in a public park |
| | S4 | A group of people from [country] playing soccer |
| | S5 | A group of people playing soccer in a public park |
| Working from home on a laptop | W3 | A person from [country] working from home and looking intently at their laptop screen |
| | W4 | A person from [country] working from home |
| | W5 | A person working from home and looking intently at their laptop screen |

**Table 2.** Additional prompts variations used for alignment scoring

The analysis also included three variations of the original prompts. The first variation removed the generative instructions such as “generate an image of” for cooking and “generate a realistic photograph-like image as if it was taken by a photographer of” for soccer, creating more concise prompts focused on content rather than image generation instructions. These prompts are presented in Table 3 and are denoted by appending “3” to each code (i.e., C3, S3, and W3). The second variation (C4, S4, J4, D4, and W4 in Table 3) experiments with removing specific words from the prompts. This included removing location specification information from the end of the prompts (e.g., “in their home” for cooking, “in a public park” for soccer, and “at a coffeeshop” for coffee).

For the remaining two scenarios, the modifications removed “A group of” from the jogging scenario and “and looking intently at their laptop screen” from the final activity. Overall, this variation examined the effect of removing selected prompt components from the original prompts. The third variation (C5, S5, J5, D5, and W5 in Table 3) created “generic” versions by removing country references entirely (e.g., “generate an image of a couple cooking a meal in their home”). These generic prompts were evaluated against all images in their respective scenarios to assess how nationality specification affects alignment scores.

In summary, alignment scores based on CLIP, ALIGN, GPT-4.1 mini were calculated for each image five separate times per activity and model, corresponding to original prompts (Table 1), revised prompt when DALL-E 3 was used, and the three prompt variations outlined in Table 3. Images were then categorized based on the labeling process described in section 4.1., which identified the presence of traditional attire and outfit impracticality for the soccer and jogging activities. This categorization allowed for comparison between two distinct groups: images showing traditional/impractical attire versus those without such representations. For each prompt variation and scenario, the average alignment scores were calculated and compared between these two groups. For example, cooking scenario images labeled as showing traditional attire were analyzed as one group, with their alignment scores compared against the group of images where people were not wearing traditional clothing. This comparative analysis was repeated for all prompt variations across the five activities and the two models.

To determine whether the observed differences in alignment scores were statistically significant, unpaired two-sample t-tests were conducted between the positive and negative image groups for each combination of text input and activity. Because many groups did not exhibit equal variance, Welch's *t*-test was used to assess the relationships between the result sets. The null hypothesis here is that there is no statistically significant difference between groups. For instance, when analyzing images of people playing soccer in impractical attire, the null hypothesis stated there would be no statistically significant difference in alignment scores based on CLIP when calculated using the text input "a group of people from [country] playing soccer" compared to images without impractical attire. This analysis was also completed for the results generated for ALIGN and GPT-4.1 mini. Finally, Hedges' g (Welch adjustment) was used to determine the effect size, which can be used to identify the size of differences between groups.

### 4.3 RQ3: Examining Revised Prompts

The final research question examined the revised prompts generated by OpenAI for input prompts, to examine if they potentially influence how nationalities are represented in generated images. These revised prompts appear to be accessible only when using the API for image generation, and it is unclear whether they can be retrieved through the standard user interface. This research question investigated how these revised prompts influenced the generated images. To accomplish this, the five prompt subsets (one for each scenario) and their corresponding revised versions were processed independently. The first step in this analysis was to identify frequently used terms, with a focus on the word "traditional" that might be systematically added to revised prompts. Then, each revised prompt was segmented into sentences using NLTK (Bird, 2006).

To facilitate semantic comparison between sentences using cosine similarity, sentence embeddings were then generated for each sentence in both scenario corpora using Sentence-BERT (SBERT) (Reimers & Gurevych, 2019) and the "all-MiniLM-L6-v2" model. K-means clustering was applied to these sentence embeddings to identify groups of semantically similar sentences, with the number of clusters selected to match the average number of sentences in the revised prompts. Each resulting cluster was qualitatively analyzed to identify its predominant theme or topic. To characterize the distinctive vocabulary of each cluster, two approaches were followed: simple frequency counting to identify common terms, and a modified TF-IDF implementation that highlighted words both frequently used within a specific cluster and rarely appearing in other clusters.

### 4.4 Dataset

The research questions were investigated using a dataset collected specifically for this study. The dataset was based on five primary prompts corresponding to the five activities (Table 1). Images were generated using DALL-E 3 through the OpenAI official API and Gemini 3 Pro Preview through the official Google API in 2025. Table 3 presents summary statistics for the dataset. For each model, 206 images were generated, resulting in 1,030 images per model across the five scenarios (i.e., 1,030 for DALL-E 3 and 1,030 for Gemini 3), and 2,060 images in total. Because each image is associated with multiple prompts that were evaluated for alignment scores, a total of 9,270 prompts were tested. The number of DALL-E 3 prompts is larger because it includes revised prompts, which were not available through the Gemini API.

| Model | Activity | Images | Prompts |
|---|---|---|---|
| DALL-E 3 | Cooking a meal | 206 | 1,030 |
| | Drinking coffee at a coffeeshop | 206 | 1,030 |
| | Jogging | 206 | 1,030 |
| | Playing soccer | 206 | 1,030 |
| | Working from home on a laptop | 206 | 1,030 |
| Gemini 3 Pro Preview | Cooking a meal | 206 | 824 |
| | Drinking coffee at a coffeeshop | 206 | 824 |
| | Jogging | 206 | 824 |
| | Playing soccer | 206 | 824 |
| | Working from home on a laptop | 206 | 824 |
| **Totals** | | **2,060** | **9,270** |

**Table 3.** Summary of dataset collected for this study

To address RQ1, the analysis used the generated images, their "traditional" and "impractical" labels, and World Bank regional income group classifications for each country. Each image corresponds to a single country, which belongs to one of seven regions and one of four income groups. After reporting the percentages of labeled images by activity and model, as well as in aggregate across the full dataset, statistical tests were conducted to examine associations between the labels and regions, and then between the labels and income groups.

For RQ2, CLIP, ALIGN, and GPT-4.1 mini, alignment scores were computed for each image relative to its original prompts (Table 1), the revised prompts (for DALL-E 3 only), and the three prompt variations (Table 2). Statistical tests were then used to assess whether alignment scores differed between labeled subsets, such as images labeled "traditional" compared to images not labeled "traditional," across the three scoring methods. Finally, to address RQ3, the study analyzed the revised prompts automatically generated by OpenAI for each image and examined their textual content using commonly used natural language processing techniques to investigate the textual elements introduced during prompt revision.

# 5. Results

## 5.1 RQ1: Representations in Generated Images

Based on the labeling of images generated across activities and countries, distinct patterns emerged in how the two models represented individuals from different nationalities. When aggregating results across all activities and models, 28.4% of images depicted individuals wearing traditional attire. In addition, among the subset of images assessed for impractical clothing in the two athletics-related activities, 15.9% depicted individuals wearing attire that was impractical for those activities.

Both models exhibited substantial rates of traditional attire, with 31.07% of DALL-E 3 images labeled as "traditional" and 25.73% of Gemini 3 Pro Preview images labeled as such. For the "impractical" label, the corresponding rates were 13.6% for DALL-E 3 and 18.21% for Gemini 3 Pro Preview. The prevalence of traditional attire also varied by activity, with the cooking and coffee activities yielding the highest percentages of images labeled as "traditional," whereas the working activity yielded the lowest percentage, likely because it specified a single individual in the prompt. Figure 4 summarizes results across the five activities for both models, reporting outcomes for the "traditional" label (left panel) and the "impractical" label (right panel).

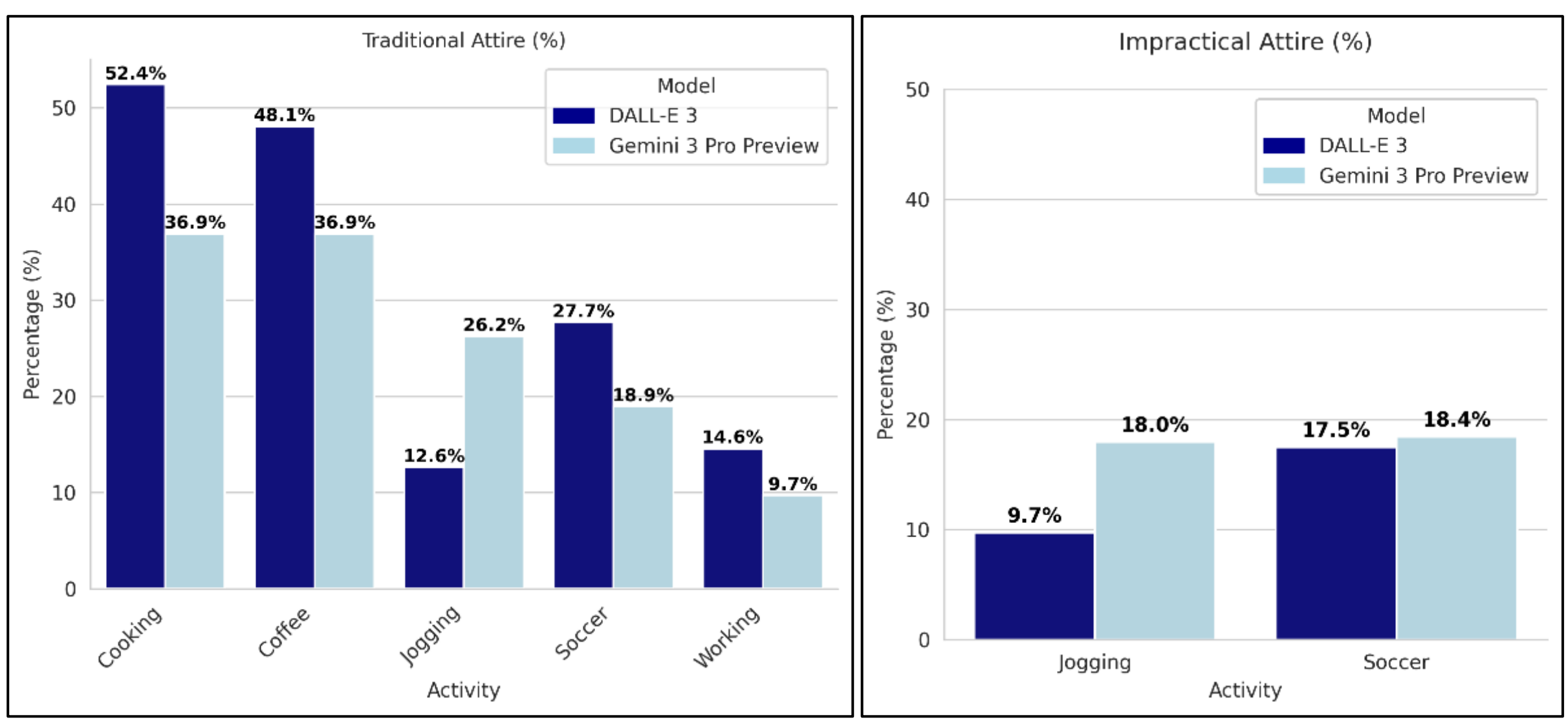

**Figure 4.** Percentages of images labeled “traditional” and “impractical,” by activity and model

Overall, these findings provide initial empirical evidence on how the two models respond to prompts requesting depictions of individuals from diverse nationalities engaged in common everyday activities. The analysis then stratified results by region and income groups to identify systematic patterns in representation. The regional analysis revealed that the Middle East & North Africa region consistently exhibited the highest percentages across activities and label categories. Countries in this region were most frequently depicted in impractical attire for soccer or jogging, as well as in traditional outfits while engaging in any of the five specified activities. For the 14 activity–label combinations, Middle East & North Africa (MENA) ranked first in the percentage of labeled images in 10 cases and second in the remaining four cases, behind South Asia. For example, for images generated using Gemini for the cooking activity, 19 of the 21 images for MENA depicted individuals in traditional attire. Across income groups, Low Income countries exhibited the highest percentages in most scenarios, often followed by Lower Middle Income countries. One country lacked World Bank income-group classification data and was excluded from the income-group analysis. Overall, these descriptive results provide initial evidence of associations between region and the labeling outcomes, as well as between income group and the labeling outcomes. However, statistical testing is required to determine whether these associations are significant.

Statistical testing using chi-square tests of independence assessed whether the observed regional and income-based differences were statistically significant. Across the 14 activity–model combinations, all tests produced large $\chi^2$ values (25 to 84.7) and moderate-to-large effect sizes, with Cramér’s V ranging from 0.34 to 0.64. However, two regions frequently produced expected cell counts below the recommended minimum of 5. To mitigate this issue, the analysis combined North America with Latin America and South Asia with East Asia & Pacific. Even after this adjustment, eight of the 14 tests still had at least one expected count below 5; in six of these eight tests, only a single cell fell below 5, and the minimum expected count across all eight was 2.04. Accordingly, results should be interpreted cautiously when the minimum expected count is below 5. Expected cell counts below 5 also occurred for the income-group analyses; therefore, these results (reported in Table 4) should likewise be interpreted with caution.

| Activity | Model | Label | Region (chi-squared test) | | | Combined Regions (chi-squared test) | | | Income Group (chi-squared test) | | |
|---|---|---|---|---|---|---|---|---|---|---|---|
| | | | $\chi^2$ | P | V | $\chi^2$ | P | V | $\chi^2$ | P | V |
| Cooking | DALL-E 3 | Traditional | 47.1* | < .001 | 0.47 | 43.6 | < .001 | 0.46 | 21.6 | < .001 | 0.32 |
| | Gemini 3 Pro | Traditional | 77.8* | < .001 | 0.61 | 70.1 | < .001 | 0.58 | 65.8 | < .001 | 0.56 |
| Drinking Coffee | DALL-E 3 | Traditional | 40.4* | < .001 | 0.44 | 37.1 | < .001 | 0.42 | 41.6 | < .001 | 0.45 |
| | Gemini 3 Pro | Traditional | 84.7* | < .001 | 0.64 | 68.1 | < .001 | 0.57 | 47.4 | < .001 | 0.48 |
| Jogging | DALL-E 3 | Traditional | 47.4* | < .001 | 0.48 | 46.8 | < .001 | 0.47 | 1.32* | 0.72 | 0.08 |
| | Gemini 3 Pro | Traditional | 70.1* | < .001 | 0.58 | 60.9* | < .001 | 0.54 | 17.7 | < .001 | 0.29 |
| | DALL-E 3 | Impractical | 37.5* | < .001 | 0.42 | 36.7* | < .001 | 0.42 | 2.17* | 0.53 | 0.10 |
| | Gemini 3 Pro | Impractical | 40.9* | < .001 | 0.44 | 28* | < .001 | 0.36 | 11.3* | 0.0098 | 0.23 |
| Playing Soccer | DALL-E 3 | Traditional | 42.5* | < .001 | 0.45 | 42.3* | < .001 | 0.45 | 8.53 | 0.03 | 0.20 |
| | Gemini 3 Pro | Traditional | 39.4* | < .001 | 0.43 | 33.6* | < .001 | 0.40 | 14.4* | 0.0024 | 0.26 |
| | DALL-E 3 | Impractical | 25* | < .001 | 0.34 | 24.9 | < .001 | 0.34 | 0.98* | 0.80 | 0.06 |
| | Gemini 3 Pro | Impractical | 39.4* | < .001 | 0.43 | 32.6* | < .001 | 0.39 | 16.4* | < .001 | 0.28 |
| Working from Home | DALL-E 3 | Traditional | 34.9* | < .001 | 0.41 | 39.6* | < .001 | 0.43 | 1.34* | 0.71 | 0.08 |
| | Gemini 3 Pro | Traditional | 39.7* | < .001 | 0.43 | 33.8* | < .001 | 0.40 | 17.9* | < .001 | 0.29 |

**Table 4.** Results for region and income groups. Stars next to $\chi^2$ indicate at least one excepted value was below 5. V here refers to Cramer's V

For the 14 tests using the combined regional categories, the analysis identified statistically significant associations between region and labels, with $\chi^2$ values ranging from 24.9 to 70.1, all p-values below 0.001, and moderate effect sizes (Cramér's V = 0.34–0.58). Full results are provided in Table 4. Results for income groups were less consistent, and some activity–model combinations did not exhibit strong associations. Finally, the expected count issues were addressed further by aggregating results across models and activities (Table 5). To provide a more holistic view, results were grouped and aggregated by activity and model. For example, results from the two models for the cooking activity were combined, and the same tests examining associations with region and income group were conducted. In addition, results were aggregated by model (e.g., all DALL-E 3 results for the "traditional" label and, separately, for the "impractical" label) as well as by label across models and activities. This produced 13 additional tested subsets, all listed and detailed in Table 5. These analyses provide a comprehensive overview of the results and the associations between the aggregated subsets and both region and income group. Most notably, when combining all 10 sets of results across the two models and five activities, 28.4% of images depicted people dressed in traditional attire and strong associations were observed with both region ($\chi^2$ = 386.1, $p < .001$, Cramér's V = 0.43) and income group ($\chi^2$ = 171.8, $p < .001$, Cramér's V = 0.28). By region, Middle East & North Africa had the highest proportion of images labeled "traditional" (70.4%), followed by South Asia (65.0%). By income group, Low Income (53.3%) and Lower Middle Income (41.6%) ranked highest. Overall, the association was moderate-to-large for region and small-to-moderate for income group.

These results provide stronger evidence of associations between countries' regions and income groups and how people from those countries are depicted in generated images. Across the aggregated subsets using the combined regional categories, all tests indicated statistically significant associations with region, with moderate-to-large effect sizes. For income groups, all p-values were below 0.01 except for one subset. Overall, these results suggest that the observed patterns are more likely to generalize across a broad set of activities rather than being driven by only one or two scenarios.

| Group/Activity | Label | N images | Labeled Positively | Region (chi-squared test) | | | Combined Regions (chi-squared test) | | | Income Group (chi-squared test) | | |
|---|---|---|---|---|---|---|---|---|---|---|---|---|
| | | | | $\chi^2$ | P | V | $\chi^2$ | P | V | $\chi^2$ | P | V |
| All Activities and Models | Traditional | 2060 | 585 (28.4%) | 386.1 | < .001 | 0.43 | 355.9 | < .001 | 0.41 | 171.8 | < .001 | 0.28 |
| | Impractical | 824 | 131 (15.9%) | 124.2* | < .001 | 0.38 | 110.3 | < .001 | 0.36 | 24.2 | < .001 | 0.17 |
| DALL-E 3 Activities | Traditional | 1030 | 320 (31.07%) | 144.0* | < .001 | 0.37 | 139.8 | < .001 | 0.36 | 53.1 | < .001 | 0.22 |
| | Impractical | 412 | 56 (13.6%) | 58.2* | < .001 | 0.37 | 57.7 | < .001 | 0.37 | 2.7 | 0.4376 | 0.08 |
| Gemini 3 Pro Activities | Traditional | 1030 | 265 (25.73%) | 264.7* | < .001 | 0.50 | 228.3 | < .001 | 0.47 | 130.8 | < .001 | 0.35 |
| | Impractical | 412 | 75 (18.21%) | 76.9* | < .001 | 0.43 | 57.7 | < .001 | 0.37 | 27 | < .001 | 0.25 |
| Cooking | Traditional | 412 | 184 (44.67%) | 110.4* | < .001 | 0.51 | 100.1 | < .001 | 0.49 | 78.1 | < .001 | 0.43 |
| Drinking Coffee | Traditional | 412 | 175 (42.48%) | 113.1* | < .001 | 0.52 | 98.1 | < .001 | 0.48 | 87.6 | < .001 | 0.46 |
| Jogging | Traditional | 412 | 80 (19.42%) | 109.3* | < .001 | 0.51 | 101 | < .001 | 0.49 | 15.6 | 0.0013 | 0.19 |
| | Impractical | 412 | 57 (13.84%) | 68.2* | < .001 | 0.40 | 56.7 | < .001 | 0.37 | 12.5 | 0.0058 | 0.17 |
| Playing Soccer | Traditional | 412 | 96 (23.31%) | 73.3* | < .001 | 0.41 | 71.2 | < .001 | 0.41 | 21.4 | < .001 | 0.22 |
| | Impractical | 412 | 74 (17.97%) | 59.7* | < .001 | 0.42 | 55.8 | < .001 | 0.36 | 12 | 0.0072 | 0.17 |
| Working from home | Traditional | 412 | 50 (12.14%) | 71.0* | < .001 | 0.38 | 70.2 | < .001 | 0.41 | 11.6 | 0.0086 | 0.16 |

**Table 5.** Results for region and income groups, aggregated by activity and model, and overall

## 5.2 RQ2: CLIP, ALIGN, and GPT-4.1 mini Efficacy

The second research question examined the alignment values computed by CLIP, ALIGN, and GPT-4.1 mini for labeled images. For each image generated by DALL-E 3 and Gemini 3 Pro Preview, alignment scores were computed using all three methods. These scores were then aggregated by activity (cooking, coffee, jogging, soccer, and working), label (traditional and impractical), and model (DALL-E 3 and Gemini 3 Pro). Results are shown in Figure 5, with each subfigure reporting outcomes for one label and one scoring method. For all five activities, scores were computed for all generated images and multiplied by 100 to facilitate easier comparison.

The scores represent average alignment scores between each image set and the original prompt used to generate the images. For example, for the prompt "a candid photograph of a group of people from [country] jogging," the mean CLIP score was 34.08 for images labeled as traditional, compared to 32.70 for images not labeled as such. The corresponding ALIGN scores were 27.71 and 25.50, respectively, while GPT-4.1 mini yielded scores of 77.81 and 69.11. It is important to note that GPT-4.1 mini does not rely on cosine similarity to assess alignment. Instead, it generates a score based on its internal reasoning. Consequently, the analysis did not aim to directly compare scores across evaluation methods. Rather, it examines whether each method consistently differentiates between positively and negatively labeled image sets across activities, prompts, and image generation models.

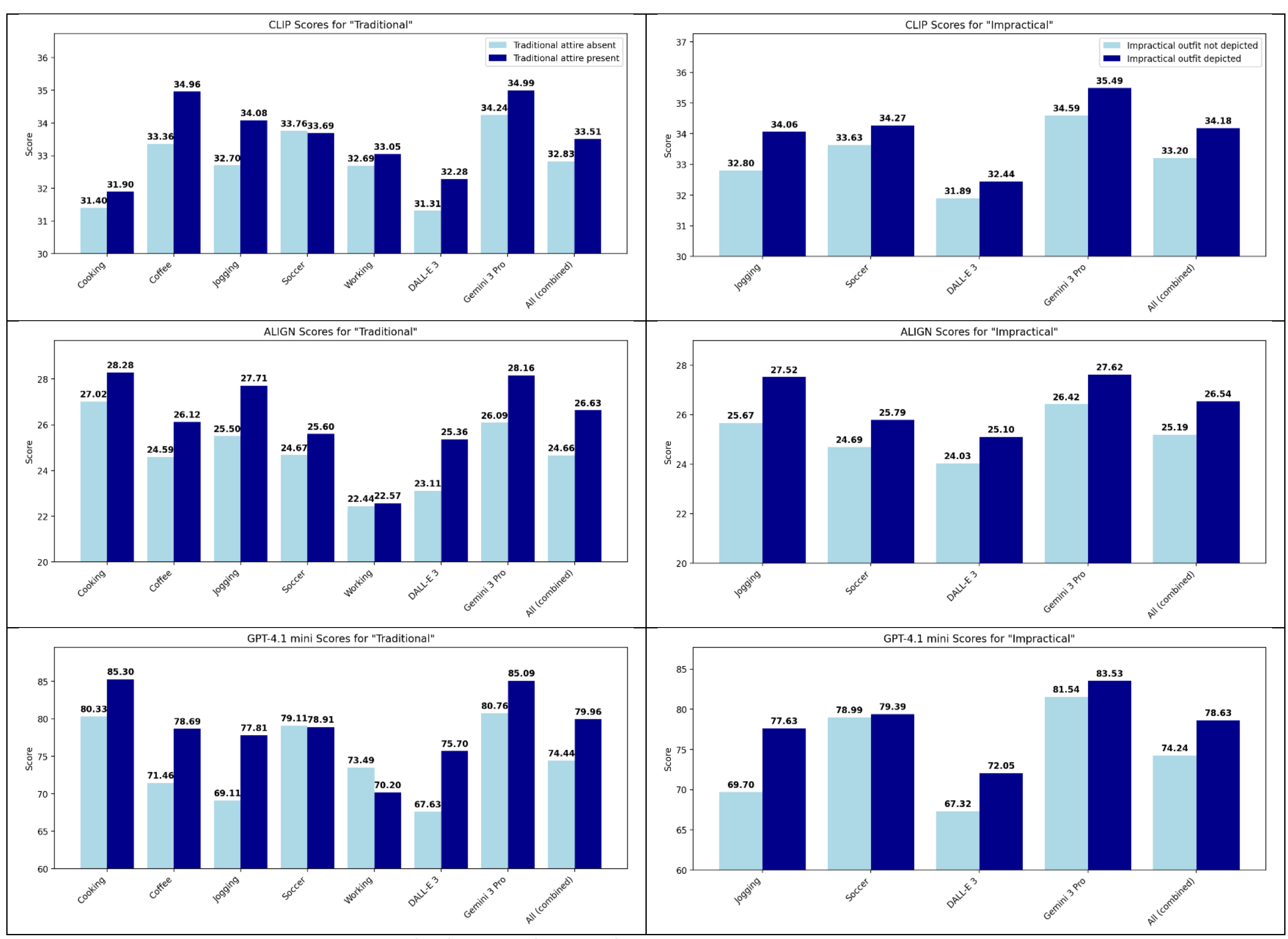

**Figure 5.** Alignments results for CLIP, ALIGN, and GPT-4.1 mini. For better readiability, the y-axis ranges have been changed

For the standard prompts used to generate the images (see Table 1 for the full prompts), an unexpected pattern emerged. Across all three scoring methods, alignment scores were consistently higher for images featuring individuals in traditional clothing compared to those without it. This pattern is also held when aggregating results across all activities for both DALL-E 3 and Gemini 3 Pro Preview. For DALL-E 3, the average scores were 31.31 vs. 32.28 (CLIP), 23.11 vs. 25.36 (ALIGN), and 67.63 vs. 75.70 (GPT-4.1 mini).

For Gemini 3 Pro Preview, the corresponding values were 34.24 vs. 34.99 (CLIP), 26.09 vs. 28.16 (ALIGN), and 80.76 vs. 85.09 (GPT-4.1 mini). Finally, when combining image sets across both DALL-E 3 and Gemini 3 Pro for all activities, images labeled as traditional received higher scores according to CLIP (32.83 vs. 33.51), ALIGN (24.66 vs. 26.63), and GPT-4.1 mini (74.44 vs. 79.96). This suggests the three scoring methods perceive a stronger prompt-image alignment when traditional clothing is present, despite the prompt not explicitly requesting traditional attire.

When examining activities individually, image sets labeled as “traditional” generally received higher alignment scores across the three methods (Figure 5). Two exceptions emerged. First, for the soccer activity, CLIP and GPT-4.1 mini produced nearly identical scores for images labeled as traditional and non-traditional. A likely explanation is that soccer images were labeled as "traditional" even when only one individual in the image wore traditional attire, whereas in other activities (e.g., cooking and drinking coffee), all individuals depicted were more consistently portrayed in traditional attire. Second, for the "working from home" activity, GPT-4.1 mini produced higher scores for images not labeled as traditional. Many of the generated "working from home" images were tightly cropped and showed only the upper body, making elements of traditional attire subtle or partially obscured. It is possible that prompts explicitly requesting full-body depictions would reduce this discrepancy. Finally, both soccer and working had fewer images labeled as either traditional or impractical than the other activities, which may also have contributed to these differences.

The same trend was observed for images labeled as impractical: positively labeled sets consistently received higher alignment scores across activities, generating models, and aggregated results. In other words, images depicting individuals wearing clothing impractical for jogging or playing soccer tended to have higher alignment with the original prompts than images that did not include such depictions. When combining results across both models and the two relevant activities, the aggregated scores further reflected this pattern for CLIP, ALIGN, and GPT-4.1 mini. Overall, these findings provide initial evidence that commonly used prompt–image alignment methods may systematically assign higher alignment to stereotyped or contextually inappropriate depictions, offering insight into how such methods implicitly weight nationality cues and clothing attributes when evaluating relevance.

The findings displayed on Figure 5 are descriptive, and statistical testing is needed to determine whether the observed differences are statistically significant. To determine whether differences between labeled image sets across activities, generative models (DALL-E 3 and Gemini 3 Pro), and alignment scoring methods (CLIP, ALIGN, and GPT-4.1 mini) were statistically significant, Welch's $t$-tests were conducted, and effect sizes were calculated using Hedge's g. For each test set, scores for images labeled as traditional or impractical were compared against scores for images not labeled as such. Results are reported in Table 6. For the traditional label, aggregating by generative model showed statistically significant differences for both DALL-E 3 and Gemini 3 Pro, with moderate effect sizes for CLIP and GPT-4.1 mini and large effect sizes for ALIGN. For the impractical label, differences were also statistically significant, but effect sizes were generally smaller for both models.

When combining all results across the five activities and both models, statistically significant differences were observed for all six test sets (label × alignment method). However, effect sizes varied and were generally on the lower end, as measured by Hedges' g: small (0.24 for traditional × CLIP; 0.25 for impractical × GPT-4.1 mini), small-to-moderate (0.33 for traditional × GPT-4.1 mini; 0.39 for impractical × CLIP), and moderate-to-large (0.49 for traditional × ALIGN; 0.61 for impractical × ALIGN). Overall, these results indicate consistent statistical differences between labeled image sets in alignment scores computed by CLIP, ALIGN, and GPT-4.1 mini, although the magnitude of the differences varies by label and scoring method. The analysis also examined revised prompts and their associated alignment scores. Table 7 reports these results, showing alignment scores between revised prompts and labeled DALL-E 3 images.

| Label | Activity/ Subset | CLIP | | | ALIGN | | | GPT-4.1 mini | | |
|---|---|---|---|---|---|---|---|---|---|---|
| | | T Test | P-Value | Effect Size | T Test | P-Value | Effect Size | T Test | P-Value | Effect Size |
| Traditional | Cooking | 2.42 | 0.016* | 0.24 | 5.83 | *** | 0.57 | 4.82 | *** | 0.48 |
| | Coffee | 5.14 | *** | 0.52 | 4.3 | *** | 0.43 | 3.92 | *** | 0.39 |
| | Jogging | 4.52 | *** | 0.57 | 5.81 | *** | 0.7 | 4.57 | *** | 0.46 |
| | Soccer | -0.19 | 0.85 | -0.03 | 3.86 | *** | 0.45 | -0.11 | 0.91 | -0.02 |
| | Working | 0.66 | 0.5 | 0.13 | 0.24 | 0.81 | 0.05 | -0.89 | 0.38 | -0.22 |
| | DALL-E 3 | 5.39 | *** | 0.39 | 9.72 | *** | 0.71 | 5.97 | *** | 0.41 |
| | Gemini 3 Pro | 4.1 | *** | 0.32 | 12.3 | *** | 0.79 | 7.17 | *** | 0.44 |
| | All (Combined) | 4.68 | *** | 0.24 | 12.13 | *** | 0.61 | 6.79 | *** | 0.33 |
| Impractical | Jogging | 3.58 | *** | 0.52 | 4.25 | *** | 0.58 | 3.72 | *** | 0.42 |
| | Soccer | 1.68 | 0.09 | 0.25 | 4.34 | *** | 0.54 | 0.2 | 0.84 | 0.03 |
| | DALL-E 3 | 1.55 | 0.12 | 0.25 | 3.33 | ** | 0.46 | 1.69 | 0.09 | 0.23 |
| | Gemini 3 Pro | 3.05 | ** | 0.44 | 3.88 | *** | 0.45 | 1.77 | 0.07 | 0.18 |
| | All (Combined) | 3.68 | *** | 0.39 | 5.37 | *** | 0.49 | 2.94 | ** | 0.25 |

**Table 6.** Original prompts test results for alignment scores identified using CLIP, ALIGN, and GPT-4.1 mini
* $p < 0.05$, ** $p < 0.01$, *** $p < 0.001$

| Label | Activity/ Subset | CLIP | | | ALIGN | | | GPT-4.1 mini | | |
|---|---|---|---|---|---|---|---|---|---|---|
| | | T Test | P-Value | Effect Size | T Test | P-Value | Effect Size | T Test | P-Value | Effect Size |
| Traditional | Cooking | 5.98 | *** | 0.82 | 5.19 | *** | 0.71 | 4.27 | *** | 0.61 |
| | Coffee | 2.66 | ** | 0.38 | 2.45 | * | 0.34 | 2.55 | * | 0.36 |
| | Jogging | 1.52 | 0.14 | 0.37 | 2.52 | * | 0.58 | 0.24 | 0.81 | 0.07 |
| | Soccer | 0.13 | 0.9 | 0.03 | 1.15 | 0.25 | 0.2 | 2.03 | * | 0.3 |
| | Working | 0.24 | 0.81 | 0.05 | 0.82 | 0.42 | 0.15 | 3.36 | ** | 0.54 |
| | All (Combined) | 6.66 | *** | 0.49 | 6.81 | *** | 0.47 | 6.2 | *** | 0.41 |
| Impractical | Jogging | 2.2 | * | 0.56 | 3.27 | ** | 0.82 | 2.09 | * | 0.47 |
| | Soccer | 0.67 | 0.5 | 0.13 | 0.96 | 0.34 | 0.19 | 1.03 | 0.30 | 0.19 |
| | All (Combined) | 1.77 | 0.08 | 0.28 | 2.98 | ** | 0.47 | 2.09 | * | 0.29 |

**Table 7.** Revised prompts test results for alignment scores identified using CLIP, ALIGN, and GPT-4.1 mini
* $p < 0.05$, ** $p < 0.01$, *** $p < 0.001$

Overall, the aggregated results remained statistically significant for revised prompts, indicating that the model-generated rewritten prompts preserve the same pattern observed with the original prompts: images labeled positively (traditional or impractical) continue to receive higher alignment scores than negatively labeled images. At the activity level, the soccer and "working from home" scenarios showed little to no difference when scored using CLIP and ALIGN, but small-to-moderate differences when scored using GPT-4.1 mini. When comparing effect sizes for images scored against original prompts versus revised prompts, effect sizes were broadly similar. This suggests that although revised prompts may contribute to the issues identified in this paper, they do not mitigate them. A potential mitigation strategy is therefore to develop revised prompts that more explicitly guide image generation models toward depictions that avoid the issues discussed in this paper.

| Label | Prompt # | Activity/ Subset | CLIP | | | ALIGN | | | GPT-4.1 mini | | |
|---|---|---|---|---|---|---|---|---|---|---|---|
| | | | T Test | P-Value | Effect Size | T Test | P-Value | Effect Size | T Test | P-Value | Effect Size |
| Traditional | 3 | DALL-E 3 | 8.2 | *** | 0.61 | 3.81 | *** | 0.27 | 7.17 | *** | 0.49 |
| | | Gemini 3 Pro | 9.67 | *** | 0.69 | 11.04 | *** | 0.69 | 8.6 | *** | 0.55 |
| | | All | 9.05 | *** | 0.47 | 5.15 | *** | 0.26 | 8.68 | *** | 0.42 |
| | 4 | DALL-E 3 | 11.19 | *** | 0.84 | 9.22 | *** | 0.63 | 11.92 | *** | 0.79 |
| | | Gemini 3 Pro | 6.72 | *** | 0.49 | 10.88 | *** | 0.64 | 11.44 | *** | 0.69 |
| | | All | 10.24 | *** | 0.52 | 8.85 | *** | 0.42 | 13.75 | *** | 0.63 |
| | 5 | DALL-E 3 | 2.82 | ** | 0.21 | -7.31 | *** | -0.49 | -1.97 | 0.050 | -0.14 |
| | | Gemini 3 Pro | 3.08 | ** | 0.22 | -0.97 | 0.33 | -0.08 | -2.03 | * | -0.14 |
| | | All | 4.1 | *** | 0.22 | -6.55 | *** | -0.34 | -2.72 | ** | -0.14 |
| Impractical | 3 | DALL-E 3 | 0.73 | 0.46 | 0.13 | 0.54 | 0.59 | 0.08 | 1.23 | 0.22 | 0.19 |
| | | Gemini 3 Pro | 4.7 | *** | 0.68 | 3.91 | *** | 0.52 | 2.07 | * | 0.21 |
| | | All | 3.7 | *** | 0.41 | 3.17 | ** | 0.33 | 2.61 | ** | 0.24 |
| | 4 | DALL-E 3 | 1.36 | 0.17 | 0.25 | 1.3 | 0.19 | 0.2 | 4.11 | *** | 0.54 |
| | | Gemini 3 Pro | 3.41 | *** | 0.49 | 3.99 | *** | 0.52 | 4.67 | *** | 0.47 |
| | | All | 3.69 | *** | 0.4 | 3.76 | *** | 0.38 | 6.42 | *** | 0.51 |
| | 5 | DALL-E 3 | -2.93 | ** | -0.52 | -5.52 | *** | -0.7 | -3.72 | *** | -0.83 |
| | | Gemini 3 Pro | 0.73 | 0.47 | 0.11 | -3.21 | ** | -0.45 | -2.62 | * | -0.32 |
| | | All | -1.24 | 0.22 | -0.15 | -5.57 | *** | -0.52 | -4.78 | *** | -0.53 |

**Table 8.** Prompts variations test results for alignment scores Identified using CLIP, ALIGN, and GPT-4.1 mini
* $p < 0.05$, ** $p < 0.01$, *** $p < 0.001$

A final analysis for RQ2 examined the three prompt variations (Table 2 provides the full text for each variation by activity). The third prompt variation for both labels removed visual descriptors from the input prompts. The fourth variation further shortened the prompts by removing additional contextual phrases (e.g., "in their home" for the cooking activity and "in a public park" for the soccer activity). The fifth variation removed country references entirely (e.g., "A group of people playing soccer in a public park"). Results are reported in Table 8.

For the traditional label, the third and fourth prompt variations produced patterns similar to those observed with the original prompts. For the impractical label, results for the "impractical × DALL-E 3" subset differed when using the third prompt variation. However, it is important to note that the impractical analyses are based on only two activities; accordingly, these findings should be interpreted cautiously.

For the fifth prompt variation, which removed country names from the prompts, a notable pattern emerged. For the traditional label, both ALIGN and GPT-4.1 mini produced average alignment scores that were consistently higher for images not labeled as traditional. Put differently, once country references were removed, these methods no longer assigned higher alignment to images depicting traditional attire. This suggests that, for ALIGN and GPT-4.1 mini, the presence of traditional clothing may be perceived as less compatible with prompts that specify only the activity (e.g., cooking at home) without an explicit country cue. The corresponding differences were moderate in magnitude for traditional × ALIGN (g = −0.34), impractical × ALIGN (g = −0.52), and impractical × GPT-4.1 mini (g = −0.53). In contrast, the effect was negligible for traditional × GPT-4.1 mini (g = −0.14). This pattern did not hold for CLIP, which exhibited inconsistent results across DALL-E 3 and Gemini 3 Pro and across the two label categories.

Overall, these findings suggest caution when using CLIP, ALIGN, and GPT-4.1 mini in this context for prompt-image alignment, particularly when evaluating images containing cultural elements or nationality-specific representations. Additionally, despite its recent release, GPT-4.1 mini did not demonstrate a significant improvement over CLIP and ALIGN in terms of alignment performance within the specific context of this study. Finally, these results highlight the complexity of the issues explored here and motivate extending the analysis to additional models and scenarios.

### 5.3 RQ3: Revised Prompts Analysis

The final research question investigated the revised prompts generated by OpenAI for the standard prompts across countries. Accordingly, RQ3 focused only on the subset images generated using DALL-E 3. This analysis revealed important indicators of how revised prompts may have influenced the T2I model to generate some of the features identified in previous sections. Despite having the same number of revised prompts, the average sentence count per revised prompt varied by activity, averaging 3.86 sentences per prompt overall.

An initial exploratory analysis processed all revised prompts generated for each of the five activity subsets to identify the most frequently used words. As expected, the most common terms included words derived directly from the input prompts, such as “coffee” and “group” for the coffee scenario and “meal” and “cooking” for the cooking activity. More notably, the revised prompts also introduced words not explicitly present in the original prompts, including “traditional” and “diverse.” In addition, revised prompts in multiple scenarios included nationality and ethnicity-specific descriptors including “African,” “Caucasian,” and “Hispanic.” Overall, the revised prompts frequently specified demographic characteristics across multiple dimensions, including gender, race, ethnicity, and age.

While many terms commonly used in revised prompts could reasonably be derived from the original prompts, it was notable that the term “traditional” appeared frequently in revised prompts for some activities. Because a primary focus of this study is labeling images based on the presence of traditional attire, this term was examined in more detail. Specifically, all revised prompts for each activity were processed to count how often the word “traditional” was added. The term appeared in 88.46% of revised prompts for the cooking activity. The percentage was also relatively high for the coffee activity, where 31.5% of revised prompts included the term. In contrast, revised prompts for jogging (10.10%), soccer (4.33%), and working (0.97%) contained the term far less frequently. Notably, the two activities with the highest rates of “traditional” in revised prompts, cooking and drinking coffee, also had the highest percentages of images labeled as traditional. This frequent inclusion of “traditional” in the cooking and coffee prompts may help explain why these two activity categories contained higher rates of traditional attire than other subsets.

This pattern suggests that revised prompts may contribute to the representation patterns observed in generated images. To examine this possibility more systematically, two sets of revised prompts were constructed for each label. Specifically, all revised prompts associated with images labeled as “traditional” were compiled into one set, while revised prompts associated with images not labeled as “traditional” were compiled into a second set. The prevalence of the word “traditional” was then measured within each subset.

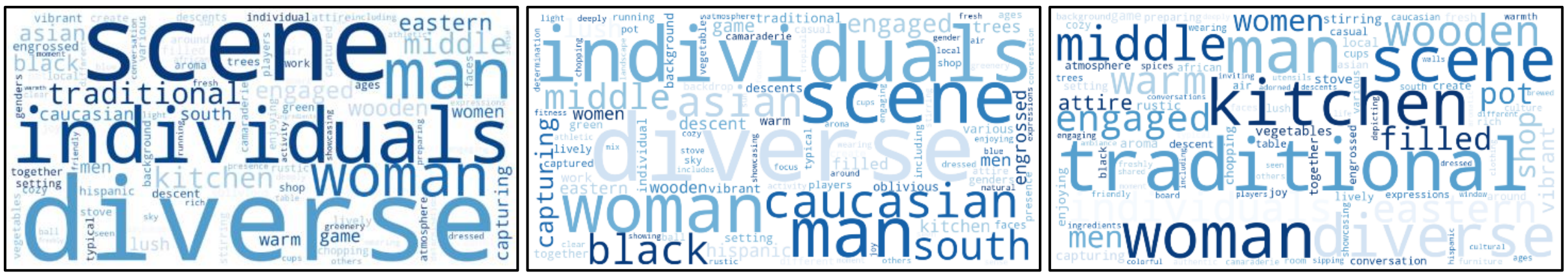

**Figure 6.** Word clouds of commonly used words that appear in revised prompts but not in the corresponding input prompts. Left: all revised prompts. Middle: revised prompts associated with images not labeled as "traditional." Right: revised prompts associated with images labeled as "traditional"

The results indicate that 50.3% of revised prompts linked to images labeled as "traditional" included the word "traditional," compared to 16.6% of revised prompts linked to images not labeled as "traditional." A similar pattern was observed when the dataset was stratified by the "impractical" label. In this analysis, 23.2% of revised prompts associated with images labeled as "impractical" contained the word "traditional," whereas only 4.7% of revised prompts associated with images not labeled as "impractical" contained the term. These results provide additional evidence that prompt rewriting may contribute to the introduction of specific cultural elements into generated images.

All revised prompts were then compiled into one large set. Then, all words from all individual revised prompts were processed and their frequencies were counted. The top words list contained mostly words that appeared in the original input prompts. To better understand what actually get added to revised prompts, a version of the top words list was created by removing all words that appeared in the original prompts used to generate the images. In this list, and using the raw frequencies of words, the top list included in the order of appearances: diverse, scene, individuals, man, woman, traditional, middle, kitchen, engaged, and Asian.

Two additional lists were created: (a) revised prompts associated with images not labeled as "traditional," and (b) revised prompts associated with images labeled as "traditional." Using word cloud visualizations, Figure 6 presents the most common words across the three subsets. As shown in the figure, the term "traditional" appears frequently in the subset of revised prompts associated with images labeled as "traditional" (right panel of Figure 6). In fact, after excluding words that also appeared in the input prompts, "traditional" was the most frequent term in this subset.

For deeper semantic analysis, all sentences from each activity corpus were embedded using SBERT (all-MiniLM-L6-v2 model) and clustered into five distinct clusters. This cluster count was selected to approximate the average sentence count per revised prompt (3.86). Five clusters, rather than four, were used to help isolate photography-related terminology, which was expected given that some original prompts requested "a candid, documentary-style photograph" or "a realistic photograph-like image as if taken by a photographer." For each cluster, two vocabulary lists were generated: one based on raw term frequency and another using a modified TF-IDF implementation that highlighted terms uniquely prevalent in a given cluster relative to others.

| Common Themes | Sample of Common Words in Theme |
|---|---|
| Background, context, and scene | Warm, scene, traditional, wooden, atmosphere, setting, sun, early, sky, park, trees, lush, home, office, environment |
| Activity, action type, and scenario details | Cooking, culinary, kitchen, coffee, engaged, brewed, jogging, running, activity, laptop, deep, captured, soccer, game, ball |
| Human representation and demographic | Couple, traditional, man, group, middle, Asian, woman, Caucasian, Black, Hispanic, diverse, people, descent, individual, South, person |
| Photography and technical specifications | Candid, photograph, image, camera, profile, presence, capturing |
| Emotional states | Love, warmth, joy, warm, laughter, relaxed, physical, expressions, determination, cheering, intently, deep, engrossed |
| Physical objects | Pot, stove, board, cup, beans, furniture, attire, ball |

**Table 9.** Qualitatively identified themes based on clustering revised prompts

Across the five activity subsets, a qualitative analysis was conducted to identify recurring themes. Although themes did not align perfectly across activities, distinct themes were nevertheless apparent when examining both the frequency-based and TF-IDF-based word lists. These themes are summarized in Table 9, along with representative terms drawn from the cluster-specific vocabularies for each activity. The complete lists of the top 15 words per cluster for each activity are provided in the appendix (Tables A.1 and A.2). In summary, revised prompts were segmented into sentences, embedded, and clustered into five groups for each activity. Two cluster-level vocabulary lists were then constructed per activity, and these lists were qualitatively reviewed to identify common thematic patterns across the five activities.

Cluster analysis revealed distinct thematic groupings in the revised prompts, each reflecting specific aspects that were also evident in the generated images. The first theme focused on scene setup, providing additional context and background specifications. Representative terms included “scene,” “traditional,” and “atmosphere.” A second theme captured activity-specific details. As expected, this theme varied substantially across activities, incorporating terminology directly related to the prompt context. For example, “coffee,” “engaged,” and “brewed” for the coffee activity, and “cheering,” “players,” and “ball” for the soccer activity.

Additional themes focused on technical or photographic attributes of the desired image, as well as emotional tone, with terms such as “joy,” “love,” and “laughter.” Another recurring theme emphasized the inclusion of specific objects, which likewise differed by activity depending on the relevant props or contextual items.

Most significantly, all five activities contained a dedicated cluster focusing on human representation. These representation-focused clusters included explicit demographic specifications based on aspects such as race and ethnicity, with terms like “genders,” “ages,” “diverse,” and “descent” appearing frequently. Specific nationality references were prominently featured, with clear nationality indicators added to numerous revised prompts. When examining revised prompts across nationalities, some included explicit guidance on how to represent diversity in the images, whereas others did not. Notably, some prompts contained explicit instructions regarding attire, such as one directive that people should be “dressed appropriately for the sport, while also reflecting their cultural context.” This particular image was subsequently labeled as both “impractical” and “traditional” in the analysis.

These findings suggest that the representation patterns identified throughout this study could potentially be addressed through strategic modifications to revised prompts. However, the effectiveness of such interventions may be limited if the underlying training datasets contain inherent representational issues. Additional research is needed to fully assess whether improving revised prompt alone can meaningfully improve representation accuracy, particularly if the foundation models have problematic representational patterns from their training data.

# 6. Discussion

## 6.1 Implications

The findings of this study demonstrate how two T2I models represent people from diverse nationalities when prompted to generate images of people engaging in common everyday activities. The results demonstrate that a substantial proportion of generated images depict individuals in traditional cultural attire and clothing impractical for the specified activities. These observations contribute to the growing body of research examining representation patterns in T2I systems, particularly in how these patterns correlate with specific geographic regions and economic indicators (Jha et al., 2024; Senthilkumar et al., 2024; Ventura et al., 2025). This research extends the existing literature, adding to discussions about the data challenges underlying these systems (Alsudais, 2025; Birhane et al., 2024; Jha et al., 2024).

While prior research has addressed related concerns (Bartl et al., 2025; Luccioni et al., 2023; Wang et al., 2022), none have specifically focused on the issue explored in this paper. As such, the first theoretical implication of this work is it brings attention to this particular challenge, which may encourage further investigation by other researchers.

By foregrounding these issues, the study may help motivate the research community to pursue solutions to the problems identified in this paper. For example, developing a benchmark dataset specifically designed to evaluate how text-to-image models represent individuals from different countries while performing various activities could guide future improvements. Such a benchmark could also incentivize model developers to incorporate these representational considerations into system design and evaluation.

Motivating model developers to address the issues identified in this paper could help future systems reduce or eliminate the observed patterns. Practical interventions should also be considered, and several directions may be fruitful for developing mitigation strategies. First, training data is frequently cited as a key driver of problematic representations. Incorporating more localized and contextually representative content, particularly for underrepresented regions, may help models develop a more accurate understanding of how people from different countries are typically depicted in everyday settings. Second, as shown in this study, revised prompts may influence how models portray certain nationalities, for example, by encouraging the inclusion of traditional attire regardless of context. Developing revised prompts strategies that promote context-appropriate and culturally accurate depictions could therefore be a valuable mitigation.

The analysis also identified specific countries that were labeled positively in most or all activities (i.e., repeatedly depicted in traditional and/or impractical attire), and it found statistically significant associations between region and income group and these labels. These findings suggest that targeted attention to the most affected regions and income groups may be particularly impactful.

Given that the set of countries examined is manageable, one practical direction is to develop nationality-specific guidance that supports context-appropriate representation. For example, because countries in the Middle East and North Africa region were frequently depicted in traditional attire regardless of activity, integrating guidance that reflects how attire is typically contextualized in daily life could help mitigate these patterns.

Other practical solutions may also exist and can be considered and tested immediately, as they may provide timely guidance. One promising direction is to leverage existing datasets. This study focused on five common activities and examined how people from various nationalities are depicted while performing them. Given the wide range of everyday activities, a practical next step is to develop a richer empirical understanding of how people in different regions typically engage in these activities and how attire varies by context.

Existing datasets, such as ActivityNet (Caba Heilbron et al., 2015), which contains videos spanning diverse activity classes, could be repurposed to support this effort. More broadly, a conceptual framework linking activity, attire, and nationality could help structure both evaluation and mitigation strategies. In addition, because in-context learning approaches have shown promise in improving model behavior, similar mechanisms could be explored to steer text-to-image systems toward more context-appropriate and culturally accurate representations.

As the findings of this study indicate, representation issues in text-to-image models are more likely to affect countries from specific regions or income groups. Therefore, another possible path towards solutions is to consider the possible strategies that countries can adopt to mitigate these issues. One such approach involves countries communicating their practical, localized needs to model developers, which may assist in addressing the challenges highlighted in this paper. Additionally, local initiatives aimed at developing customized solutions that reflect a country's unique values and characteristics could be valuable. However, such efforts would likely require substantial resources, which may not be accessible to many countries.

AI companies might also consider providing regionally customized experiences that account for users' geographic and cultural contexts. To the best of our knowledge, AI tools such as ChatGPT and Gemini currently adapt to user location when processing text-based prompts. Extending similar practices to text-to-image models could improve user experience and representation. For instance, when a user requests an image of "a couple cooking a meal in their home," the model could generate an image that reflects cultural and contextual elements relevant to the user's location. Developers of these tools also have access to a large and diverse user base across many countries. While academic research teams may encounter logistical and resource-related challenges when conducting studies that require recruiting participants from multiple countries, such processes may be more straightforward for companies. This access positions them well to systematically investigate users' representation preferences across different cultural and national contexts.

This paper also uncovered limitations in commonly used evaluation metrics, which often fail to capture the issues identified in this study. Thus, another contribution of this paper is the presentation of empirical evidence highlighting the need for new evaluation metrics for text-to-image alignment that account for fairness and appropriate representation across national identities. Previous research has proposed modifications to CLIP to enhance its performance or applications (Song et al., 2025; Sun et al., 2024). Other researchers introduced alternative evaluation metrics aimed at addressing current limitations (Huang et al., 2025; Lin et al., 2024).

These efforts do not address the specific issue examined in this paper. Future research that develops evaluation metrics capable of identifying and potentially mitigating unfavorable representations of individuals from specific countries in text-to-image models would represent a valuable contribution to the field. The alignment results for the recently released GPT-4.1 mini, which demonstrated similar patterns to those observed with CLIP and ALIGN, suggest that even the most current models remain incapable of detecting the types of issues highlighted in this study.

Overall, these directions represent plausible avenues for addressing the issues identified in this paper. In summary, they include (1) developing new benchmarking datasets to evaluate how text-to-image (T2I) models represent people from different nationalities; (2) incorporating more localized and contextually representative data during model development; (3) improving revised-prompt construction to reduce systematic biases in depiction; (4) proposing new evaluation methods that overcome the limitations of existing alignment scoring methods often employed in this context; and (5) prioritizing nationalities and regions that are most consistently affected.

A more immediate and practical approach is to leverage existing resources such as ActivityNet and develop an activity–attire–nationality framework. Such a framework could then be used within an in-context learning setup to steer models toward more context-appropriate and culturally accurate representations. A final practical solution relies on incorporating country-specific inputs and on model developers considering interventions informed by their ability to access localized user bases.

The findings of this paper provide valuable insights and offer theory-relevant evidence that connects directly to prior research on how computational systems “see” and represent social groups. The results for RQ1 show that representational patterns are structured rather than random. Specifically, there is a statistically significant relationship between World Bank region and World Bank income group and the likelihood that the model generates images emphasizing “traditional” attire and stereotypical depictions, with disproportionately strong effects for Middle East & North Africa and Sub-Saharan Africa. These patterns align with Said’s theory of Orientalism, which argues that dominant representational system often represent the “non-West” using essentialized and exoticized cultural signifiers instead of contemporary outlook (Said, 1978). Recent works extend this theory through Digital Orientalism, examining how digital media and online platform represent specific regions (Alimardani & Elswah, 2021; Nguyen, 2025). Related work on data and algorithmic orientalism highlights how the design of computational systems can produce data outputs that encode representational inequities (Kotliar, 2020). Similarly, work on “Decolonial AI” argues for examining AI systems from a global perspective that accounts for harms across diverse geographic contexts, rather than focusing primarily on North American or European framings (Mohamed et al., 2020).

Building on these foundations, this paper introduces *AI Orientalism* as a generative-AI analogue of these prior accounts. It refers to a tendency for generative AI models, shaped likely by the combined influences of training data, prompt revision, and evaluation, to default to traditionalized and culturally essentializing visual cues as a shortcut for depicting certain nationalities, particularly in contexts where such cues are impractical. This tendency can systematically underrepresent contemporary appearance and everyday life for people from countries in specific regions and income groups. Additional research is needed to assess whether these patterns extend to other generative modalities, such as text-to-video systems and large language models.

The findings for RQ2 indicate that representational concerns are also likely shaped by commonly used evaluation models designed to quantify how well outputs align with input prompts. Most notably, when country names were included in prompts, CLIP, ALIGN, and GPT-4.1 mini assigned significantly higher alignment scores to images labeled as containing “traditional” attire. This suggests that traditionalized visual elements operate as a strong correlational cue for nationality terms in these models’ evaluation behavior. This pattern is consistent with recent research on shortcut learning, which shows that models may rely on highly learnable correlations that improve measured performance without capturing the intended real-world construct (Geirhos et al., 2020).

This interpretation is reinforced by the results for the fifth prompt variation: when country names were removed from prompts, images labeled as “traditional” no longer received higher alignment, and the trend reversed for ALIGN, and GPT-4.1 mini, with depictions not including traditional attire scoring higher. Taken together, these results suggest that scores from the three evaluation methods are significantly driven by nationality linked cues rather than activity relevant cues, which raises a construct validity concern. In other words, the metric employed, in this case alignment score, appears to diverge from the intended evaluative target. This aligns with emerging work emphasizing construct validity as a primary concern in AI evaluation and benchmarking (Bean et al., 2025; Freiesleben & Zezulka, 2025). The findings for RQ3 provide additional evidence that AI Orientalism may be reinforced by prompt intervention mechanisms that are possibly designed as well-intended “improvements,” but that nonetheless contribute to the representational concerns identified in this study. Specifically, the substantially higher prevalence of the word “traditional” in revised prompts associated with images labeled as “traditional,” relative to those not labeled as such, suggests that the observed representational skew is likely also shaped by the design decisions that rewrite user inputs prior to image generation.

When considering the findings from RQ1, RQ2, and RQ3, results indicate that these representational concerns plausibly operate across multiple levels of the generation pipeline, including image generation, evaluation, and prompt intervention. This interpretation aligns with established frameworks emphasizing that harms and biases can be introduced at different stages of the machine learning lifecycle, including through design choices and evaluation practices, thereby contributing to representational harms (Suresh & Guttag, 2021).

### 6.2 Limitations

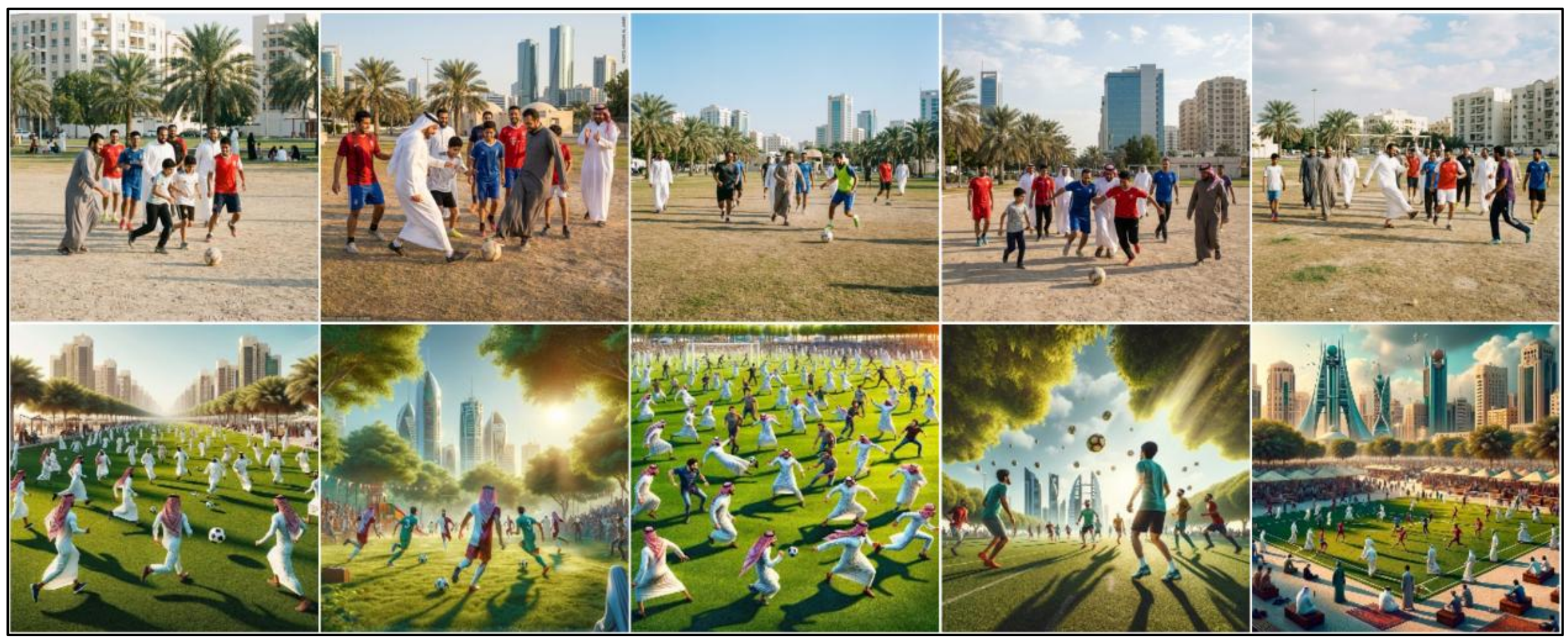

**Figure 7.** Sample images for one country and one activity, demonstrating consistent results

This research has several limitations. First, the findings presented are based on testing only two T2I models. Therefore, the findings may not be generalizable to all T2I models and could be more commonly occurring when this specific model is used. Second, only five activities were tested and thus, similarly, results may not be easily generalized. Similarly, the findings for ALIGN reflect an open-source version rather than the original ALIGN model. Thus, the findings may not be generalizable to the original ALIGN model. Third, revised prompts for the cooking and soccer scenarios were generated in early 2025, whereas revised prompts for the remaining activities were generated in late 2025. It is therefore unclear whether changes to the revised prompt generation process between these periods may have affected the observed distributions. Finally, how people from one country wish to be portrayed is not easily agreed upon and disagreements likely exist between members of the same group regarding appropriate representation (Hall, Bell, et al., 2024). Still, these limitations present opportunities for future research.

Testing a variety of T2I systems would reveal whether the representation issues identified in this study are model-specific or represent broader patterns across different systems. Such comparative analysis focusing specifically on nationality representation could establish important benchmarks for evaluating future models. Additionally, expanding the research to include additional common activities beyond the five examined in this study would likely uncover more nuanced patterns in how different activities influence representational patterns. Finally, investigating approaches to updating generated images using corrective prompts could be a viable opportunity for future research.

A final limitation concerns the fact that only one image per model and activity was generated. Put differently, because the results rely on a single image for each country, model, and activity combination, the outcomes could differ if the same prompt were rerun. However, experimentation throughout this work showed consistent results. In addition, selected country, model, activity combinations that were tagged as both traditional and impractical were tested. For example, five images were generated for one country for the soccer activity using DALL-E 3 first and then Gemini 3 Pro (Figure 7). The results showed that the generated images mostly fit the criteria (i.e., nine of the ten images were labeled as both traditional and impractical). Still, future research could examine this issue more systematically by generating multiple images for each combination and evaluating the labels more thoroughly.

# 7. Conclusion

In this paper, two common approaches to generating images were tested to assess how they represent people from various nationalities performing common activities. Results revealed that generated images frequently depicted people in traditional or impractical attire. Statistical analysis showed significant relationships between these representations and both World Bank geographic regions and income groups. These findings highlight significant issues in how these models handle requests centered around nationalities. Through multiple analytical approaches including assessing CLIP, ALIGN, GPT-4.1 mini, and revised prompts, this study demonstrates the extent of these representation issues. These findings can inform efforts to mitigate such unfavorable representations in future T2I systems. These representation problems are unlikely to improve without deliberate intervention by developers of AI tools.

Evidence from long-running online platforms such as YouTube and X indicates that safeguards and quality controls are often uneven across languages and regions, with non-English and lower-resourced contexts frequently receiving less consistent coverage. For example, one study analyzed YouTube search results for content in the Ethiopian language and found that such searches often returned problematic content, in contrast to searches in English (Nigatu & Raji, 2024). Therefore, unless developers of AI tools allocate dedicated resources and maintain ongoing monitoring and mitigation efforts, the types of issues identified in this study are likely to persist over time. A practical implication of this work is its highlighting of how text-to-image models depict individuals from specific nationalities. This insight may inform future model development and contribute to the creation of more accurate and equitable representations.

Building on the empirical findings of this study, which indicate that representational concerns can arise at multiple stages of the model pipeline (generation, evaluation, and prompt rewriting), this paper introduces the concept of AI Orientalism. This contribution extends prior theoretical work on Orientalism, digital Orientalism, and data/algorithmic Orientalism, and it complements recent research on bias in AI, construct validity in AI evaluation, and shortcut learning.

Ultimately, the paper identifies one class of issues that may contribute to representational harms affecting certain nationalities. At the same time, it is important to recognize that these models provide substantial benefits across a wide range of use cases and application contexts. The purpose of this work is to document and clarify these representational patterns and to support ongoing efforts to evaluate and mitigate them in future systems.

## Appendix

```
You are an Image–Caption Alignment Judge.
### Goal
Given **one image** and **the following caption**:{caption_text}", you must:
1. Compute an **Alignment Score** ∈ [-1, 1] (floating-point, any precision).
   - **-1.0** → completely unrelated or contradictory
   - ** 0.0** → neutral / no discernible relation
   - ** 1.0** → perfectly matches in all salient aspects
   The score should approximate a cosine similarity between the visual scene and the text.
2. Write a **Concise Justification** (1–3 well-formed sentences) that cites concrete visual/textual evidence for *why* you chose that score.
### Reasoning Steps (think but only output final answer)
1. **Identify** prominent objects, actions, attributes in the image.
2. **Parse** key facts in the caption.
3. **Compare** each element, note matches, partials, mismatches.
4. **Aggregate** into a single scalar score.
5. **Craft** a brief justification citing the strongest evidence.
### Output Format (JSON-like list, no additional keys)
```

**Figure A.1.** Prompt used to Instruct GPT-4.1 mini to Generate Alignment Scores

| | Cluster | | | | |
|---|---|---|---|---|---|
| **Activity** | **1** | **2** | **3** | **4** | **5** |
| Cooking | ['pot', 'woman', 'man', 'chopping', 'vegetables', 'stirring', 'stove', 'wooden', 'board', 'fresh', 'cutting', 'large', 'simmering', 'stew', 'traditional'] | ['warm', 'room', 'window', 'home', 'scene', 'warmth', 'filled', 'atmosphere', 'traditional', 'love', 'cultural', 'glow', 'aroma', 'typical', 'sense'] | ['cooking', 'meal', 'together', 'traditional', 'preparing', 'joy', 'engaged', 'process', 'shared', 'home', 'expressions', 'faces', 'love', 'culinary', 'kitchen'] | ['kitchen', 'couple', 'home', 'meal', 'man', 'woman', 'image', 'traditional', 'cooking', 'middle', 'preparing', 'descent', 'together', 'Caucasian', 'create'] | ['kitchen', 'ingredients', 'utensils', 'wooden', 'spices', 'filled', 'traditional', 'rustic', 'cooking', 'pots', 'warm', 'various', 'vegetables', 'home', 'fresh'] |
| Coffee | ['coffee', 'conversation', 'cups', 'engaged', 'conversations', 'engrossed', 'sipping', 'man', 'others', 'woman', 'enjoying', 'brewed', 'laughter', 'cup', 'warm'] | ['group', 'middle', 'Asian', 'south', 'Caucasian', 'individuals', 'man', 'descents', 'woman', 'black', 'women', 'men', 'eastern', 'diverse', 'Hispanic'] | ['warm', 'scene', 'wooden', 'atmosphere', 'setting', 'relaxed', 'laughter', 'rustic', 'casual', 'soft', 'conversations', 'tables', 'hanging', 'people', 'capturing'] | ['coffee', 'group', 'candid', 'diverse', 'individuals', 'coffeeshop', 'shop', 'photograph', 'local', 'people', 'enjoying', 'image', 'capturing', 'scene', 'men'] | ['coffee', 'aroma', 'wooden', 'shop', 'air', 'warm', 'rustic', 'freshly', 'brewed', 'furniture', 'atmosphere', 'cozy', 'coffeeshop', 'beans', 'tables'] |
| Jogging | ['lush', 'background', 'landscape', 'backdrop', 'greenery', 'tropical', 'trees', 'typical', 'clear', 'blue', 'natural', 'green', 'sky', 'beautiful', 'setting'] | ['group', 'men', 'women', 'Caucasian', 'middle', 'black', 'Hispanic', 'south', 'Asian', 'eastern', 'diverse', 'ages', 'individuals', 'people', 'descents'] | ['jogging', 'group', 'diverse', 'candid', 'individuals', 'photograph', 'people', 'together', 'engaged', 'capturing', 'image', 'activity', 'running', 'attire', 'men'] | ['sun', 'scene', 'sky', 'morning', 'casting', 'shadows', 'sunlight', 'early', 'backdrop', 'trees', 'light', 'setting', 'park', 'long', 'clear'] | ['faces', 'camaraderie', 'determination', 'expressions', 'fitness', 'athletic', 'attire', 'physical', 'joy', 'activity', 'exercise', 'group', 'dressed', 'wear', 'vibrant'] |
| Soccer | ['park', 'trees', 'lush', 'green', 'sky', 'clear', 'blue', 'grass', 'public', 'scene', 'greenery', 'filled', 'background', 'vibrant', 'sun'] | ['image', 'photographer', 'photograph', 'professional', 'realistic', 'taken', 'create', 'generate', 'captured', 'capturing', 'like', 'skilled', 'detailed', 'scene', 'quality'] | ['group', 'diverse', 'men', 'women', 'game', 'black', 'soccer', 'people', 'players', 'descent', 'individuals', 'Caucasian', 'Asian', 'Hispanic', 'ages'] | ['image', 'group', 'soccer', 'park', 'public', 'diverse', 'generate', 'realistic', 'photograph', 'game', 'individuals', 'like', 'people', 'men', 'women'] | ['soccer', 'ball', 'game', 'players', 'park', 'action', 'scene', 'mid', 'others', 'air', 'cheering', 'joy', 'expressions', 'energy', 'camaraderie'] |
| Working | ['laptop', 'screen', 'profile', 'camera', 'intently', 'thought', 'oblivious', 'completely', 'deep', 'engrossed', 'presence', 'person', 'looking', 'captured', 'individual'] | ['style', 'home', 'documentary', 'working', 'candid', 'photograph', 'Hispanic', 'man', 'individual', 'person', 'Caucasian', 'image', 'male', 'south', 'work'] | ['camera', 'presence', 'oblivious', 'completely', 'screen', 'laptop', 'work', 'capturing', 'moment', 'focus', 'engrossed', 'focused', 'intently', 'attention', 'profile'] | ['style', 'home', 'documentary', 'candid', 'working', 'photograph', 'person', 'individual', 'middle', 'black', 'image', 'eastern', 'Asian', 'work', 'south'] | ['work', 'home', 'concentration', 'environment', 'setting', 'dedication', 'office', 'focus', 'engrossed', 'room', 'atmosphere', 'image', 'profile', 'scene', 'workspace'] |

**Table A.1.** Vocabulary for clusters using basic frequency counts

| | **Cluster** | | | | |
|---|---|---|---|---|---|
| **Activity** | **1** | **2** | **3** | **4** | **5** |
| Cooking | ['chopping', 'stirring', 'vegetables', 'pot', 'board', 'woman', 'stove', 'man', 'cutting', 'fresh', 'wooden', 'stew', 'simmering', 'face', 'stirs'] | ['window', 'room', 'warm', 'cultural', 'atmosphere', 'warmth', 'domestic', 'glow', 'scene', 'background', 'textiles', 'sun', 'sense', 'love', 'life'] | ['together', 'joy', 'expressions', 'shared', 'cooking', 'engaged', 'culinary', 'faces', 'meal', 'preparing', 'process', 'side', 'traditional', 'show', 'conversation'] | ['couple', 'image', 'home', 'meal', 'kitchen', 'create', 'middle', 'descent', 'black', 'man', 'woman', 'Caucasian', 'Hispanic', 'asian', 'traditional'] | ['kitchen', 'utensils', 'ingredients', 'spices', 'pots', 'pans', 'elements', 'countertop', 'spread', 'shelves', 'scattered', 'various', 'like', 'filled', 'decor'] |
| Coffee | ['coffee', 'engaged', 'conversation', 'cups', 'others', 'sipping', 'cup', 'laughing', 'steaming', 'conversations', 'engrossed', 'holding', 'man', 'brewed', 'hands'] | ['caucasian', 'descents', 'mix', 'asian', 'south', 'man', 'ages', 'young', 'woman', 'middle', 'black', 'women', 'aged', 'Hispanic', 'men'] | ['glow', 'hanging', 'lamps', 'relaxed', 'spirit', 'casting', 'warm', 'soft', 'atmosphere', 'window', 'every', 'tranquil', 'working', 'filtering', 'plush'] | ['diverse', 'candid', 'group', 'individuals', 'coffeeshop', 'photograph', 'coffee', 'local', 'shop', 'capturing', 'enjoying', 'people', 'image', 'snapshot', 'descent'] | ['coffee', 'aroma', 'air', 'beans', 'freshly', 'wooden', 'brewed', 'furniture', 'rustic', 'charm', 'lighting', 'shop', 'warm', 'filled', 'tables'] |
| Jogging | ['lush', 'landscape', 'background', 'greenery', 'vegetation', 'trees', 'tropical', 'clear', 'architecture', 'backdrop', 'typical', 'blue', 'scenery', 'distant', 'skies'] | ['men', 'women', 'group', 'Hispanic', 'ages', 'Caucasian', 'south', 'man', 'genders', 'black', 'middle', 'Asian', 'descents', 'eastern', 'woman'] | ['jogging', 'diverse', 'candid', 'group', 'individuals', 'photograph', 'engaged', 'people', 'together', 'session', 'capturing', 'descent', 'jog', 'image', 'engaging'] | ['shadows', 'casting', 'sun', 'long', 'sunlight', 'soft', 'early', 'sky', 'light', 'morning', 'horizon', 'hues', 'day', 'shining', 'orange'] | ['faces', 'camaraderie', 'expressions', 'fitness', 'determination', 'physical', 'exertion', 'joy', 'athletic', 'exercise', 'health', 'shared', 'commitment', 'sweat', 'attire'] |
| Soccer | ['trees', 'sky', 'park', 'lush', 'blue', 'green', 'clear', 'overhead', 'grass', 'greenery', 'casting', 'sun', 'tall', 'shadows', 'background'] | ['photographer', 'professional', 'taken', 'skilled', 'captured', 'photograph', 'realism', 'mimic', 'image', 'resembles', 'though', 'realistic', 'depth', 'real', 'create'] | ['black', 'women', 'men', 'caucasian', 'Hispanic', 'Asian', 'descent', 'south', 'group', 'represent', 'ages', 'eastern', 'genders', 'man', 'diverse'] | ['group', 'image', 'generate', 'photograph', 'public', 'soccer', 'diverse', 'realistic', 'park', 'individuals', 'men', 'women', 'like', 'engaging', 'game'] | ['ball', 'players', 'soccer', 'air', 'mid', 'action', 'game', 'joy', 'faces', 'cheering', 'chasing', 'others', 'determination', 'expressions', 'camaraderie'] |
| Working | ['screen', 'profile', 'laptop', 'intently', 'camera', 'oblivious', 'thought', 'looking', 'presence', 'completely', 'captured', 'deep', 'engrossed', 'gazing', 'seen'] | ['Hispanic', 'documentary', 'style', 'home', 'working', 'photograph', 'man', 'candid', 'male', 'islands', 'Caucasian', 'photographic', 'portrayal', 'dominican', 'American'] | ['camera', 'presence', 'attention', 'seem', 'oblivious', 'moment', 'completely', 'appear', 'focus', 'capturing', 'intense', 'focused', 'pay', 'directed', 'experience'] | ['documentary', 'style', 'photograph', 'home', 'candid', 'working', 'middle', 'eastern', 'black', 'Asian', 'south', 'person', 'individual', 'image', 'female'] | ['room', 'light', 'around', 'atmosphere', 'modern', 'homely', 'elements', 'environment', 'personal', 'workspace', 'concentration', 'subtly', 'typical', 'expression', 'sense'] |

**Table A.2.** Vocabulary for clusters using TF-IDF